\documentclass[10pt,twocolumn,letterpaper]{article}

\usepackage[pagenumbers]{cvpr} 

\definecolor{cvprblue}{rgb}{0.21,0.49,0.74}
\usepackage[pagebackref,breaklinks,colorlinks,allcolors=cvprblue]{hyperref}
\usepackage{cuted}    
\usepackage{multirow}
\usepackage[normalem]{ulem}
\usepackage[table,xcdraw]{xcolor}
\useunder{\uline}{\ul}{}
\usepackage{wrapfig}
\usepackage{graphicx}
\usepackage[accsupp]{axessibility}  

\definecolor{alizarin}{rgb}{0.82, 0.1, 0.26}
\hypersetup{
  urlcolor   = alizarin!85!black,
}

\def\paperID{} 
\def\confName{CVPR}
\def\confYear{2026}

\title{FireScope: Wildfire Risk Raster Prediction with a Chain-of-Thought Oracle}

\author{
Mario Markov$^{1, \dagger}$ \,\,\,
Stefan Maria Ailuro$^{1}$ \,\,\,
Luc Van Gool$^{1}$ \,\,\,
Konrad Schindler$^{2}$ \,\,\,
Danda Pani Paudel$^{1}$
\\
$^{1}$ INSAIT, Sofia University ``St. Kliment Ohridski'' $\quad$
$^{2}$ ETH Zurich\\ 
{$^{\dagger}$ \tt\small mario.markov@insait.ai}\\
[0.4em]
\url{https://firescope.ai/research}
}

\begin{document}
\maketitle
\begin{abstract}
Predicting wildfire risk is a reasoning-intensive spatial problem that requires the integration of visual, climatic, and geographic factors to infer continuous risk maps. Existing methods lack the causal reasoning and multimodal understanding required for reliable generalization. We introduce \textbf{FireScope-Bench}, a large-scale dataset and benchmark that couples Sentinel-2 imagery and climate data with expert-defined risk rasters across the USA, and real wildfire events in Europe for cross-continental evaluation. Building on this dataset, we propose \textbf{FireScope}, a VLM-based reasoning-to-generation framework that learns from both reinforcement learning and visual supervision to predict risk rasters with complementary reasoning traces. When trained in the USA and tested in Europe, \textbf{FireScope} achieves substantial performance gains, while expert feedback and automated analysis confirm that its reasoning traces are faithful and semantically meaningful. Our findings demonstrate that reasoning can ground raster prediction models, improving both generalization and interpretability. To our knowledge, this is the first framework to (1) demonstrate that language-based reasoning can improve generalization in visual generation, (2) propose a high-resolution wildfire risk model that can be applied across continents, and (3) enable systematic studies of robust cross-continental generalization for multimodal fire risk models. We believe that \textbf{FireScope-Bench} has the potential to serve as a foundation for advancing reasoning-driven, interpretable and generalizable spatial modeling.


\end{abstract}

\section{Introduction}
\label{sec:intro}

\begin{figure}[t]
    \includegraphics[width=.98\linewidth]{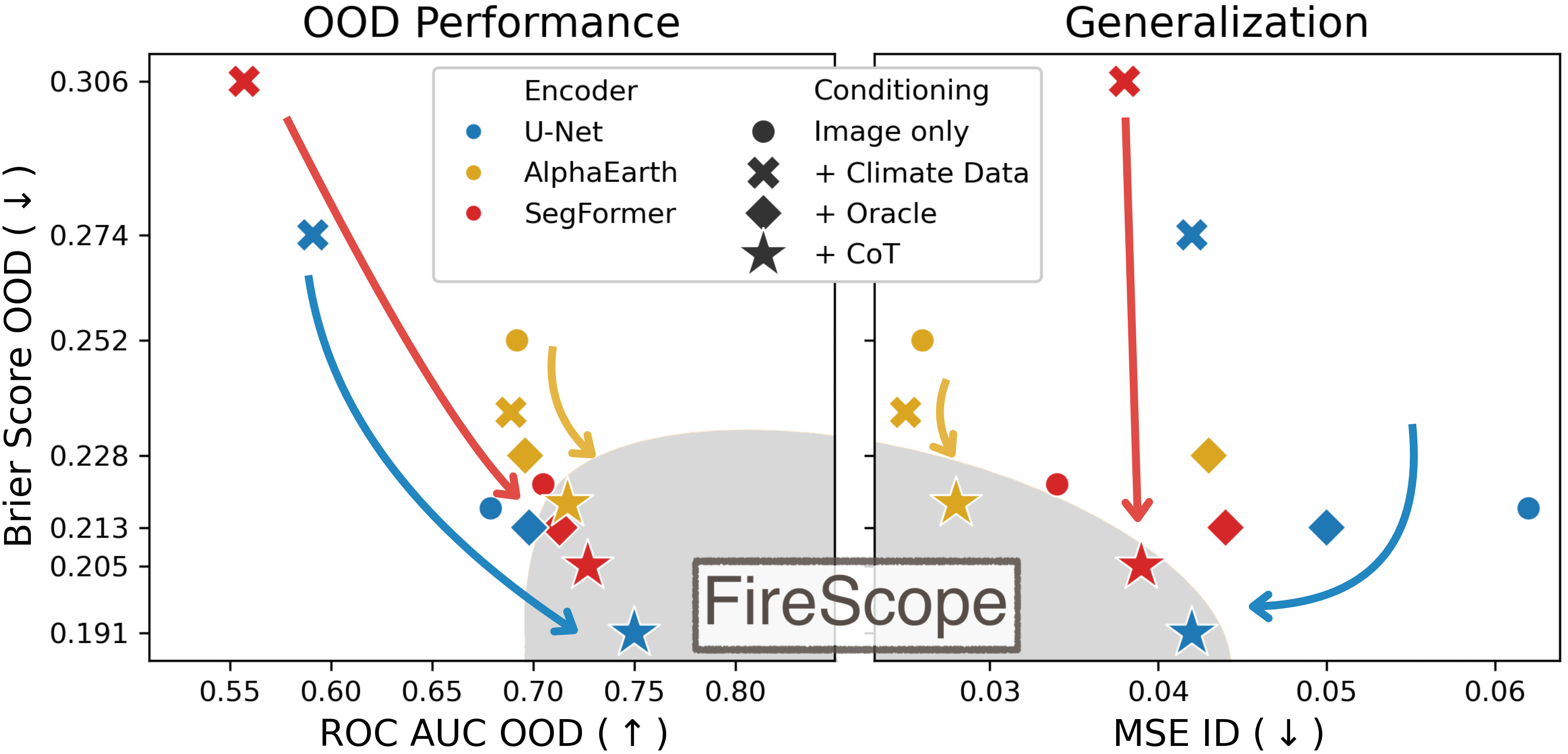}\vspace{-0.3cm}
    \setlength{\belowcaptionskip}{-5pt}
    \caption{Effects of conditioning and reasoning on wildfire risk prediction. Left: out-of-distribution (OOD) performance. Right: Generalization trade-off between OOD and in-distribution (ID) performance. More conditioning towards reasoning (→ CoT) results in better OOD generalization while remaining robust ID.
    \vspace{-2mm}
    }
    \label{fig:storyplot}
\end{figure}
\begin{figure}[t]
    \centering
    \includegraphics[width=.9\linewidth]{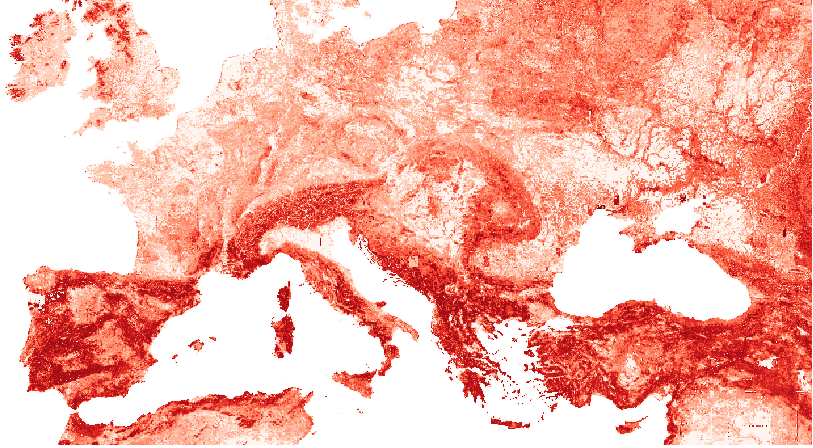}\vspace{-0.3cm}
    \setlength{\belowcaptionskip}{-5pt}
    \caption{FireScope-predicted wildfire risk across Europe for 2026. Darker red indicates higher risk. See Figure~\ref{fig:bench_fig2} for distribution of ground truth wildfires in previous years.
    \vspace{-5mm}
    }
    \label{fig:europe_risk_map}
\end{figure}

\begin{figure*}[t]
    \centering
    \includegraphics[width=\linewidth]{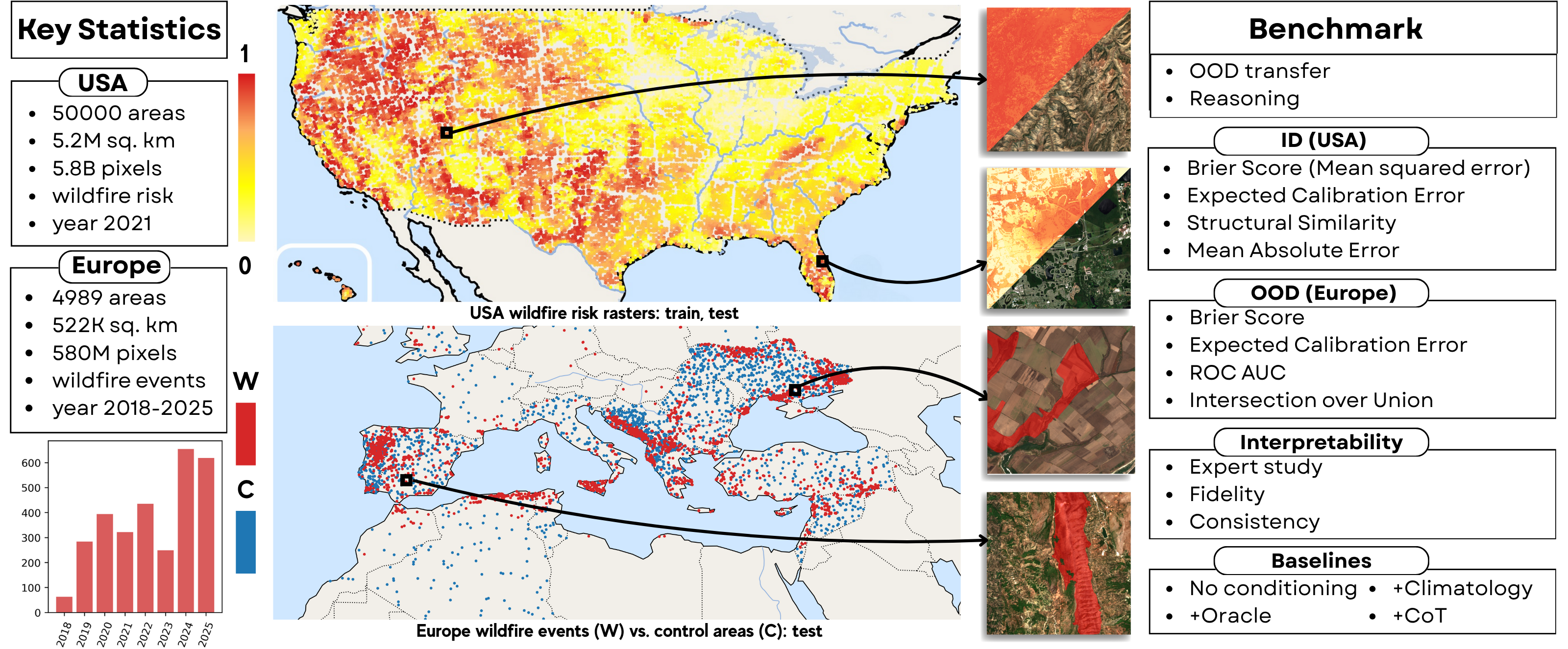}\vspace{-15pt}
    \setlength{\belowcaptionskip}{-15pt}
    \captionof{figure}{
    \emph{FireScope-Bench overview}. A large-scale multimodal benchmark combining satellite imagery, climate data, and \emph{expert-defined} risk maps over the U.S. and Europe. It enables training on USA data and testing across Europe on real wildfire events to evaluate model generalization and reasoning in wildfire risk prediction. The benchmark includes metrics for accuracy, calibration, and interpretability.
    }
    \label{fig:bench_fig2}
\end{figure*}

Accurately assessing wildfire risk is a critical challenge for both environmental science and public safety. It requires synthesizing a wide range of factors -- vegetation, topography, climate interactions, human activity, and more~\cite{DiGiuseppe2025NatComms, McNorton2024PoF} -- that extend far beyond localized image features alone. Unlike the visual task of fire detection or segmentation~\cite{arlovic2025evaluation}, wildfire risk estimation is a multimodal reasoning problem: it demands integrating visual appearance with non-visual, causal drivers to infer an abstract, spatially structured quantity -- the continuous risk raster that expresses the fine-grained wildfire risk across an area.

Despite its significance, wildfire risk prediction remains largely unexplored within the vision community. Existing datasets and models mostly address \emph{observational} tasks -- such as fire detection, segmentation, or spread estimation~\cite{arlovic2025evaluation,UncAwareDL2025,Shadrin2024SpreadDL,Buch2023SMLFire} -- rather than the more abstract goal of estimating a continuous \emph{risk field}. Traditional approaches in environmental modeling rely on physics-based indices like the Fire Weather Index (FWI)~\cite{VanWagner1987_FWI} or hybrid climate models~\cite{McNorton2024PoF,DiGiuseppe2025NatComms,TorresVazquez2025HybridSeasonal,Bohlmann2024NGRFWI}, which capture coarse-scale meteorological variability but omit high-resolution visual and geographic context. In contrast, data-driven computer vision methods typically learn from imagery alone, limiting their ability to generalize across biomes and continents~\cite{beery2018recognition,koh2021wilds}. There currently exists no unified benchmark that integrates multi-source data -- imagery, climate, and topography -- nor a framework that spans the full modeling spectrum from pixel-level visual understanding to multimodal reasoning over causal drivers. This absence of standardized datasets, metrics, and modeling paradigms hinders progress toward reasoning-capable spatial predictors. To address these gaps, we construct \textbf{FireScope-Bench}, a dataset \& benchmark that explicitly tests reasoning-intensive wildfire risk estimation.  We also introduce \textbf{FireScope}, a framework that couples vision-language reasoning with raster generation, improving the generalization of various baselines, as shown in Figure~\ref{fig:storyplot}.

FireScope-Bench (see Figure~\ref{fig:bench_fig2}) combines Sentinel-2 imagery, area-wide climatology, and \emph{expert-defined} long-term wildfire risk rasters across the continental United States, together with a distinct evaluation set of actual wildfire events and control regions across Europe. The benchmark is designed to test both in-distribution (USA) performance and cross-continental (Europe) generalization -- a setting where purely visual or correlation-based models often fail~\cite{beery2018recognition,koh2021wilds}. 
FireScope-Bench spans over 5.7M km² across 55K regions and 6.3B pixels. The USA partition (50K areas, 2021) supports training and calibration, while the European partition (4,989 areas, 2018–2025) provides wildfire events and control zones for evaluation. For completeness, we also include wildfire events in the USA; see the supplementary material. This large-scale, multimodal design enables systematic study of reasoning-driven wildfire risk estimation and cross-continental generalization.

Building on this dataset, we propose \textbf{FireScope}, a novel reasoning-to-generation framework for structured image prediction. We first fine-tune a vision-language model (VLM) with group relative policy optimization (GRPO) to reason about a given area and produce a scalar wildfire risk estimate summarizing both image and climate information. The resulting “Oracle” VLM then conditions a lightweight vision model with its scalar predictions through feature-wise linear modulation (FiLM)~\cite{perez2018film}. The vision model learns to generate continuous risk rasters grounded in the satellite imagery and consistent with Oracle’s judgments. In doing so, FireScope combines the interpretability and causal grounding of language-based reasoning with the spatial precision of vision-based decoding. To our knowledge, the proposed design is the first of its kind, which showcases that sequentially trained reasoning and vision modules can yield complementary benefits in language and visual spaces, within the discussed settings.

Empirically, FireScope achieves substantial out-of-distribution (OOD) gains over strong baselines. These improvements occur while remaining robust in-distribution, suggesting that explicit reasoning encourages the model to rely on complex causal, generalizable features rather than spurious correlations tied to local appearance. Within the ID settings, our trade-offs align with the state-of-the-art~\cite{zhao2025trade}, whereas we offer new evidence of better OOD generalization for the task at hand.

Beyond accuracy, FireScope provides transparency into the raster generation: its Chain-of-Thought (CoT) explanations are human-interpretable, allow domain experts to reconstruct risk judgments, and remain stable under paraphrasing yet sensitive to semantic perturbation. Together, these findings highlight reasoning as a powerful mechanism for grounding structured prediction models in generally meaningful evidence.

In summary, this work makes two key contributions:
\begin{itemize}
    \item \textbf{FireScope-Bench:} a large-scale multimodal dataset \& benchmark for wildfire risk prediction that demands complex reasoning across imagery, climate, and geography, enabling rigorous cross-continental OOD evaluations;
    \item \textbf{FireScope:} a reasoning-to-generation framework that unifies language-based reasoning and visual prediction, leading to improved generalization and interpretability.
\end{itemize}
Our approach encourages causal, generalizable feature use, providing a path toward more transparent and socially relevant vision systems for wildfire risk and beyond.

\section{Related works}\label{sec:relworks}

\textbf{Reasoning in Vision-Language Models}. Reasoning supervision has recently emerged as a powerful tool for improving generalization and interpretability in large models. In language models, CoT prompting encourages step-by-step reasoning that enhances accuracy on complex tasks~\cite{wei2022chain,kojima2022zeroshot,wang2022selfconsistency}. Similar ideas extend to multimodal settings, where VLMs generate rationales to explain visual predictions~\cite{suris2023viper,liu2023mmreact,zhang2023visualcot}. Yet most efforts target discrete QA-style reasoning; benchmarks that explicitly test reasoning in \emph{image generation/editing} are relatively rare~\cite{jin2024reasonpix2pix,lin2024vqascore,meng2024phybench,kajic2024numerical,r2i2025}, and focused on natural image generation, not rasters encoding spatially aligned, physically meaningful variables. We instead explore whether explicit reasoning can benefit \emph{raster generation}: FireScope-Bench couples multimodal inputs with reasoning-intensive rasters and real event masks, and FireScope leverages a CoT-trained VLM to guide spatial generation within this setting.

\noindent\textbf{Raster Generation}.
Raster generation tasks -- such as segmentation, depth estimation, and geospatial prediction -- require generating dense, spatially structured outputs. Classical approaches rely on fully convolutional networks and Encoder-Decoder architectures~\cite{ronneberger2015unet, chen2018deeplab, badrinarayanan2017segnet, long2015fully}. Recent advances have further expanded conditional synthesis through diffusion- and transformer-based models~\cite{rombach2022ldm,esser2021vqgan,balaji2022ediffi}. These methods typically learn direct correlations between input and output modalities rather than reasoning about latent causal structure. Differently, FireScope treats structured prediction as a reasoning-to-generation problem, leveraging the reasoning and generalization power of large pre-trained VLMs to interpret multimodal inputs and condition spatial decoders. This design combines the broad transfer capabilities of VLMs, enabling interpretable and robust out-of-distribution generalization.

\noindent\textbf{Wildfire Risk and Geospatial Modeling}.
Wildfire risk forecasting has traditionally relied on physics-based or statistical indices such as the Canadian Fire Weather Index (FWI)~\cite{VanWagner1987_FWI}, which remain operational standards but depend solely on meteorological variables. Recent efforts extend them with more predictors, hybrid dynamical-machine learning approaches~\cite{TorresVazquez2025HybridSeasonal, McNorton2024PoF, DiGiuseppe2025NatComms}, and statistical calibration~\cite{Bohlmann2024NGRFWI}, but still remain constrained by coarse spatial scales and climate-only or hand-crafted inputs. Deep learning and probabilistic models further enhance fire danger estimation~\cite{kondylatos2022wildfire,Shadrin2024SpreadDL, Buch2023SMLFire}, but most either operate at coarse grids or are fine-scale only in limited regions, and rarely produce continuous, interpretable risk {rasters}. While explainable methods improve transparency~\cite{Liu2024SHAPXAI,Xu2025ISPRS,Sengupta2025XAI}, they are largely post-hoc and with great computational overhead. In contrast, FireScope integrates multimodal factors with reasoning-grounded supervision to produce spatially resolved, interpretable risk rasters beyond purely meteorological pipelines.
\section{FireScope-Bench}
\label{sec:benchmark}

\begin{figure*}[ht]
  \centering
  \includegraphics[width=\linewidth]{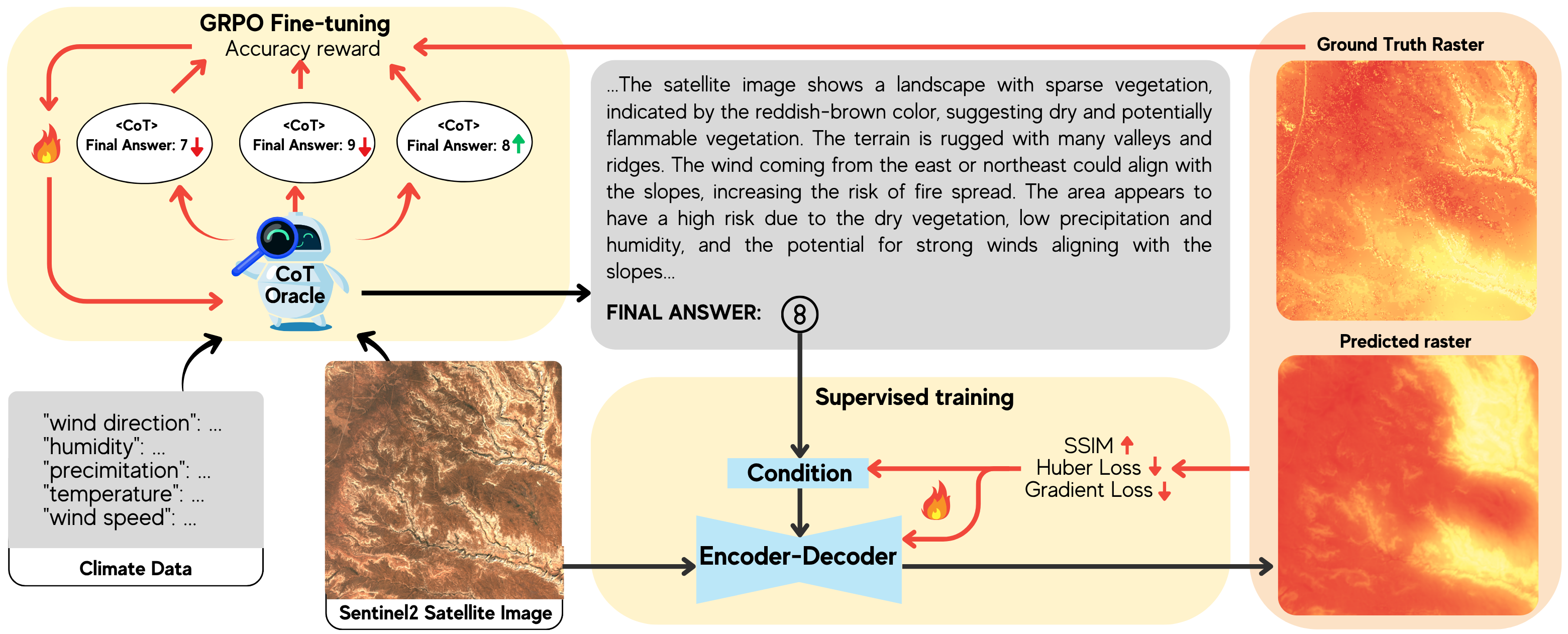}\vspace{-15pt}
  \setlength{\belowcaptionskip}{-15pt}
  \captionof{figure}{\emph{FireScope overview}. A VLM fine-tuned with GRPO learns CoT reasoning over climate and imagery to predict scalar risk (``Oracle”), which subsequently conditions Encoder–Decoder through a FiLM mechanism to generate fine-grained risk rasters, linking reasoning with spatial prediction.}
  \label{fig:short_method}
\end{figure*}

\subsection{Dataset}

\textbf{Wildfire risk rasters}. To enable high-resolution wildfire risk prediction, we curate a dataset from Wildfire Risk to Communities~\cite{wrc}, a probabilistic wildfire modeling project. It is provided in the exceptionally high resolution of 30m per pixel covering the United States.
We obtain the \emph{Risk to Potential Structures} to use as a target variable. It is a measure of the expected consequences of wildfires to potential structures, given the probability of burning and the potential fire intensity. Risk to Potential Structures integrates both wildfire likelihood and generalized consequences.

The risk raster is tiled into images of $341\times341$ pixels, approximately $100\text{km}^2$ each. We generate 40K training, 4K validation, and 4K testing samples across the mainland US and Hawaii using stratified sampling based on geographical location and risk distribution within each subset. In addition, we generate a smaller, more manageable dataset of 1K training, 100 validation, and 100 testing samples following the same procedure for faster experimentation. The tiles of the large training set are grouped into $2\times2$ supertiles to enable runtime augmentations. Finally, we normalize the values via quantile transform into the [0,1] range, turning them into a measure of relative risk. Onward, we refer to the target variable as just \emph{wildfire risk}.

\noindent\textbf{Wildfire events.} An essential part of FireScope-Bench is evaluation against real wildfire events out of the spatial distribution of the training set -- the downstream task vision models usually fail to complete~\cite{beery2018recognition,koh2021wilds}. Thus, we obtain wildfire events from the EFFIS burnt areas collection~\cite{effis, effispaper} to access the OOD transfer of the models.

We curate a dataset of 3K wildfire events in Europe, spanning from 2018 to 2025. To obtain a balanced spatial distribution, we limit the number of events in each country proportionally to the total area of its European part and filter out wildfires with areas less than five square kilometers. Additionally, 2K inland control locations with no detected wildfires are sampled proportionally to the number of wildfire events in each country. Each data sample comprises a wildfire mask within a tile of the same dimension as the risk rasters. We also include complementary events for the USA; refer to supplementary. See Figure~\ref{fig:bench_fig2} to assess the final distribution.


\noindent\textbf{Predictors}. Common wildfire prediction models require a range of variables carefully gathered by experts: vegetation, fuel content, weather, ignition factors, and potential for fire-propagation ~\cite{DiGiuseppe2025NatComms,McNorton2024PoF}, which limits the scalability of these approaches. An intelligent system should be able to deduce the required factors from raw data; recent developments in the earth observation support this approach~\cite{bastani2023satlaspretrainlargescaledatasetremote, brown2025alphaearthfoundationsembeddingfield}. Therefore, we limit the predictors to high-resolution \emph{satellite imagery} and \emph{climatology} variables.

For each wildfire risk raster, the corresponding optical satellite image is downloaded from the Sentinel-2 level 2A collection~\cite{sentinel} in 10m resolution, constituting images of $1024\times1024$ pixels. Regions occluded by clouds are excluded, and each final tile is combined as a mosaic of cloud-free images from the summer season, ranging from June 22 to September 22. The imagery is gathered from the year 2021 for wildfire risk rasters in the USA and control samples in Europe, and from the year preceding each event for wildfire events in Europe to assess the ability of models to predict wildfire rather than detect it. Furthermore, the predictors are completed by monthly climatology obtained from NASA POWER~\cite{power,powerdata}. The climatology is provided at 50km resolution. We include near-surface temperature, precipitation, humidity, wind speed, and wind direction for the centroid of every tile -- a vector of $\text{dim}=60$.

\subsection{Metrics}
\label{subsubsec:Evaluation_Metrics}
For In-distribution continuous risk rasters, we employ \textit{Mean Squared Error (MSE)} and \textit{Mean Absolute Error (MAE)} for per-pixel evaluation, and \textit{Structural Similarity Index (SSIM)}~\cite{Wang2004_SSIM} for assessing structural consistency. For OOD wildfire events discrimination we use probabilistic metrics: \textit{Brier score}~\cite{Brier1950}, \textit{Area Under the Receiver Operating Characteristic Curve (ROC AUC)}~\cite{Fawcett2006_ROC}, \textit{Expected Calibration Error (ECE)}~\cite{Naeini2015_ECE}; for per-pixel predictions of burnt areas we use \textit{ROC AUC} and \textit{Intersection over Union (IoU)}~\cite{Jaccard1912_IoU}. Since burnt pixels comprise less than 15\% of data, the Brier score is not used per-pixel as it misleadingly rewards models biased toward the negative class~\cite{Hoessly2025On}. 
Note that due to the stochastic nature of ignition, we treat non-burnt areas within wildfire tiles as background rather than negative samples~\cite{Marc-Andre2009}. Therefore, the maximum expected performance within individual images is limited, and predictions must be assessed based on the ability to distinguish wildfire events from control areas in addition to the quality of segmentation.

To evaluate Oracles, we substitute spatial metrics in-distribution with \textit{Quadratic Weighted Kappa (QWK)}~\cite{cohen1968weightedkappa}, as they produce ordinal prediction from 0 to 9 for the whole area, and we discretize the mean value of ground truth raster accordingly. 

Details on metrics implementations are provided in the supplementary material.
\section{FireScope Framework}
\label{sec:method}

FireScope is a two-stage reasoning-to-generation framework for modeling spatial wildfire risk from multimodal inputs (see Figure~\ref{fig:short_method}). In the first stage, we fine-tune the \emph{Oracle} -- a VLM tasked to infer the overall wildfire risk of a region through explicit reasoning. Given satellite imagery and auxiliary climate normals, the Oracle integrates latent cross-modal interactions to produce a scalar risk estimate that captures the overall risk score of the area.

During the second stage, a lightweight \emph{vision Encoder-Decoder} learns to produce a fine-grained, continuous risk raster conditioned on the Oracle’s scalar estimate. This allows FireScope to leverage the generalization of large pre-trained VLMs while retaining the spatial precision of the vision decoder. These two stages enable reasoning-grounded, transparent, and spatially mapped wildfire risk prediction.

\begin{table*}[ht]
\centering
  \caption{Models performance over FireScope-bench.  The section `wildfire events' refers to distinguishing burnt areas from control ones, while `wildfire pixels' -- fine-grained prediction. 
  The top-3 results are underscored in each column. The best ones are highlighted in bold. The exemplary output is provided for each method. The addition of a CoT Oracle results in substantial OOD gains.}
  \label{tab:models_metrics_ood}{\vspace{-0.2cm}}

\begin{subtable}[b]{0.68\linewidth}
  \resizebox{\linewidth}{!}{
\begin{tabular}{ll|ccc|cc}
\toprule
\multicolumn{2}{c|}{Method} &
  \multicolumn{3}{c|}{OOD wildfire events} &
  \multicolumn{2}{c}{OOD wildfire pixels} \\
Conditioning &
   Encoder &
  Brier $\downarrow$ &
  ROC AUC $\uparrow$ &
  ECE $\downarrow$ &
  ROC AUC $\uparrow$ &
  IoU@0.5 $\uparrow$ \\
\midrule
\multirow{3}{*}{Image only} &
  AlphaEarth &
  0.252 &
  0.692 &
  0.190 &
  0.632 &
  0.164 \\
 &
  SegFormer &
  0.222 &
  0.705 &
  0.106 &
  0.651 &
  {\ul 0.179} \\
 &
  U-Net &
  0.217 &
  0.679 &
  {\ul \textbf{0.050}} &
  0.587 &
  0.159 \\
  \midrule
\multirow{3}{*}{+ Climate Data} &
  AlphaEarth &
  0.237 &
  0.689 &
  0.162 &
  0.629 &
  0.173 \\
 &
  SegFormer &
  0.306 &
  0.557 &
  0.224 &
  0.518 &
  0.131 \\
 &
  U-Net &
  0.274 &
  0.591 &
  0.167 &
  0.559 &
  0.145 \\
  \midrule
\multirow{3}{*}{+ Oracle} &
  AlphaEarth &
  0.228 &
  0.696 &
  0.141 &
  {\ul 0.653} &
  {\ul 0.179} \\
 &
  SegFormer &
  {\ul  0.213} &
  0.713 &
  {\ul  0.074} &
  0.642 &
  0.177 \\
 &
  U-Net &
  {\ul  0.213} &
  0.698 &
  0.087 &
  {\ul 0.655} &
  {\ul 0.181} \\
  \midrule
\multirow{3}{*}{+ CoT}
 &
  AlphaEarth &
  0.218 &
  {\ul 0.717} &
  0.124 &
  {\ul 0.653} &
  0.177 \\
 &
  SegFormer &
  {\ul 0.205} &
  {\ul 0.727} &
  0.078 &
  {\ul \textbf{0.658}} &
  {\ul \textbf{0.184}} \\
 &
  U-Net &
  {\ul \textbf{0.191}} &
  {\ul \textbf{0.750}} &
  {\ul 0.068} &
  0.652 &
  0.178 \\
  \bottomrule
\end{tabular}
}
    \label{fig:short-a}
  \end{subtable}
  \hfill
  \begin{subtable}{0.29\linewidth}
  \includegraphics[width=\linewidth]{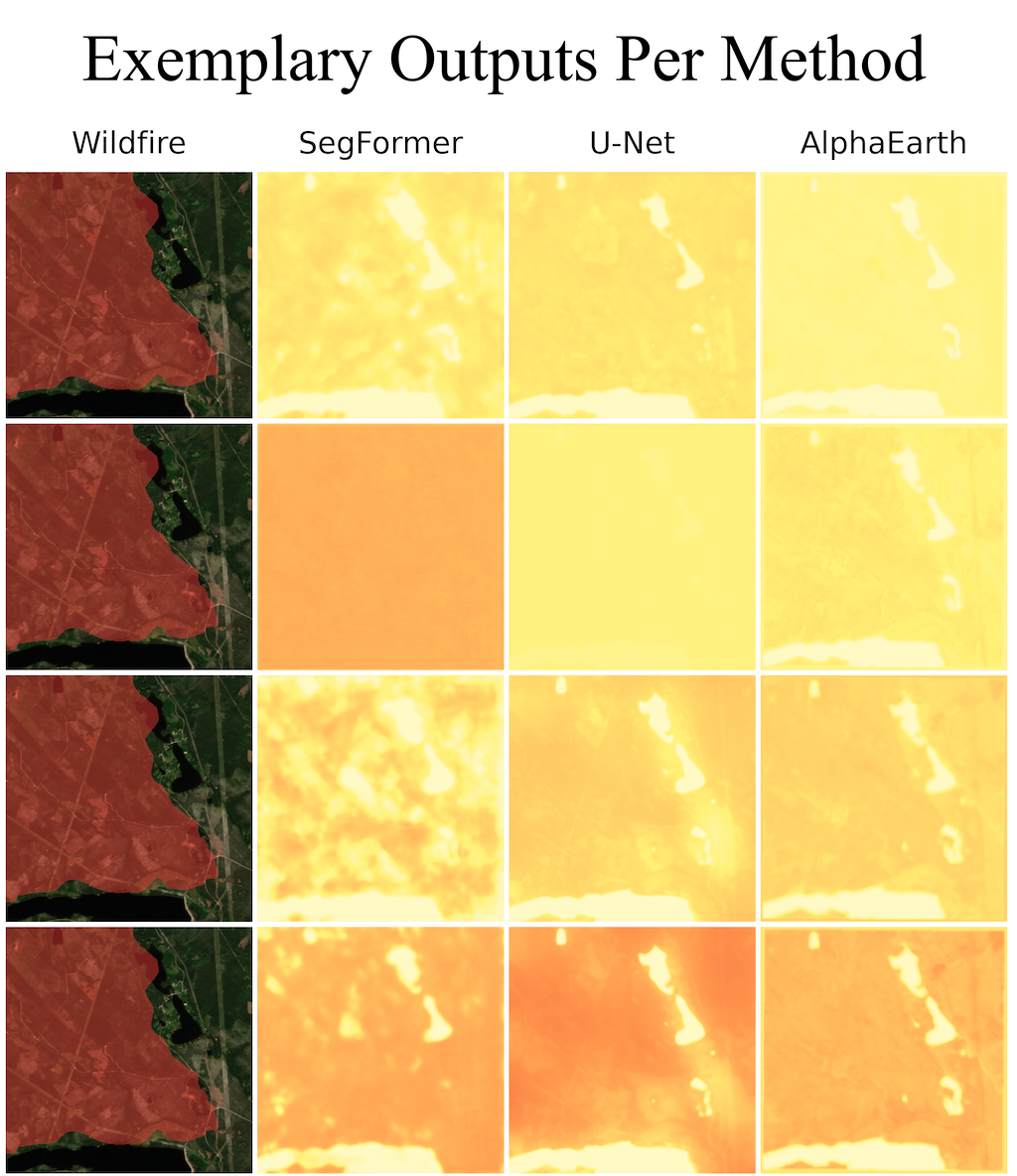}
    \label{fig:short-b}
  \end{subtable}

\vspace{-20pt}
\end{table*}

\subsection{Oracle Fine-Tuning}\label{subsec:oracle_fine_tuning}
 Whilst the ordinal labels defined in Section~\ref{subsubsec:Evaluation_Metrics} can be used to perform supervised fine-tuning on the Oracle, this would restrict it to outputting a single scalar score. To allow it to explore intermediate reasoning steps and refine its estimate iteratively, we fine-tune it with reinforcement learning (RL), which is unhindered by the lack of gradients for arbitrary-length outputs. For that, we use group-relative policy optimisation (GRPO) \cite{shao2024deepseekmath} -- a novel reinforcement learning algorithm which foregoes the need for a critic model, thus incurring much less overhead compared to other widely used RL algorithms~\cite{schulman2017ppo, haarnoja2019soft}.

\noindent\textbf{Reward Design}.
Following common practice~\cite{guo2025deepseekr1} we define our reward as the weighted sum of two components:
\begin {equation}
R = 0.9 \cdot R_{\mathrm{acc}} + 0.1 \cdot R_{\mathrm{fmt}}
\end{equation}
where $R_{\mathrm{acc}}$ rewards the \textbf{accuracy} of ordinal prediction and
$R_{\mathrm{fmt}}$ rewards correct \textbf{format}, both spanning from 0 to 1. For more details, refer to the supplementary.
To tackle the label imbalance, we employ frequency-weighted aggregation.
Despite recent advancements in rewarding the quality of intermediate outputs~\cite{zhang2025towards}, we do not explicitly steer the model's reasoning, letting it naturally evolve and adapt to the sole task of yielding high-accuracy responses. The increasingly elaborate CoTs observed during training are additional evidence that reasoning helps wildfire risk prediction.

\subsection{Vision Encoder-Decoder Training}\label{subsec:enc_dec_training}
After fine-tuning the Oracle, we deterministically generate its outputs for the training set and use them as conditioning for the vision Encoder-Decoder model via the FiLM mechanism~\cite{perez2018film}. FiLM layers are applied before each trainable block. The decoder is trained to regress a normalized risk raster $y\in[-1,1]^{341\times 341}$ by minimizing the weighted sum of three terms:
\begin{equation}
\!\underbrace{\mathcal{L}_{\text{s}l_1}(y,\hat{y})}_{\text{reconstruction}}
+
0.5\underbrace{(1-\text{SSIM}(\tilde{y},\tilde{\hat{y}}))}_{\text{structure}}
+
0.2
\underbrace{\mathcal{L}_{\ell_1}(\nabla y,\nabla\hat{y})}_{\text{edges}}
\end{equation}
where $\hat{y}$ is the model output, 
$\tilde{y}$ and $\tilde{\hat{y}}$ are matched into $[0,1]$ range.
The reconstruction term is a Smooth-$\ell_1$ loss with $\beta=1.0$. 
The Structural Similarity Index (SSIM)~\cite{Wang2004_SSIM} uses a $11\!\times\!11$ gaussian window.
The edges $\ell_1$ loss matches first-order finite differences,
encouraging sharper boundaries.

\section{Experiments} \label{sec:experiments}
For computational feasibility, unless stated otherwise, all experiments are conducted on the small training set described in Section~\ref{sec:benchmark}. The training details, hyperparameters, and the sensitivity with respect to them are reported in the supplementary material.

\subsection{Configurations}

\textbf{Oracles}. We select Qwen2.5-VL-7B-Instruct \cite{bai2025qwen2_5vl} 
as our Oracle due to its strong performance on multimodal reasoning benchmarks and efficient 7B parameter size. We train two versions: one with CoT reasoning as outlined in Section~\ref{subsec:oracle_fine_tuning}, and one trained with supervised fine-tuning without CoT. We evaluate the classification performance of three additional Oracles: \emph{GPT-5}~\cite{openai2025gpt5} -- a powerful general-purpose VLM, \emph{Fire Weather Index (FWI)}~\cite{VanWagner1987_FWI} -- a well-established meteorological-based index, and \emph{Climate MLP} -- a small multilayer perceptron over the climatology vector.

\noindent\textbf{Encoders.} We evaluate three Encoders: \emph{SegFormer} MiT-B5~\cite{xie2021segformer} -- a state-of-the-art encoder for image segmentation, \emph{AlphaEarth}~\cite{brown2025alphaearthfoundationsembeddingfield} -- a foundation model pretrained on diverse satellite tasks, and lightweight \emph{U-Net}~\cite{ronneberger2015unet} trained from scratch. To each encoder, we attach a decoder adapted to its architecture. The encoders of SegFormer and AlphaEarth are kept frozen, while the U-Net is trained entirely. For implementation details, refer to the supplementary material.

\noindent\textbf{Conditionings.} For each choice of Encoder-Decoder, we train four versions: \emph{Baseline} trained only on satellite images, \emph{Climate}-conditioned, \emph{Oracle} -- conditioning on Qwen fine-tuned without CoT, and \emph{CoT Oracle (FireScope)} -- conditioning on Qwen fine-tuned with CoT.

\begin{table}[t]
\centering
  \caption{Oracle performance and coarse-resolution baselines. The best result in each column is highlighted in bold. Reasoning significantly improves generalization.}
  {\vspace{-0.2cm}}
  \label{tab:oracle_metrics}
  \resizebox{\columnwidth}{!}{
\begin{tabular}{l|ccc|ccc}
\toprule
\multirow{2}{*}{Method}  & \multicolumn{3}{c|}{OOD wildfire events} &  \multicolumn{3}{c}{ID wildfire risk}   \\
 & Brier $\downarrow$ & ROC AUC $\uparrow$ & ECE $\downarrow$ & Brier $\downarrow$ & MAE $\downarrow$ & QWK $\uparrow$ \\
    \midrule
Climate MLP & 0.276  & 0.524 & 0.150  & \textbf{0.029} &	\textbf{0.123}	& \textbf{0.766} \\
FWI          & 0.321  & 0.551 & 0.255  & --          & --   & --    \\
GPT          & 0.281  & 0.636 & 0.229  & 0.081 &	0.225 &	0.316      \\
Qwen      & 0.225  & 0.701  & 0.134  & 0.036	 & 0.135 &	0.751      \\
CoT Qwen   & \textbf{0.196}  & \textbf{0.748}  & \textbf{0.077}   & 0.031	& 0.127 &	\textbf{0.766}         \\
    \bottomrule        
\end{tabular}
}
\vspace{-3pt}
\end{table}

\begin{table}[t]
\centering
  \caption{Models performance in-distribution. The top-3 results are underscored in each column. The best ones are highlighted in bold. While not being the best interpolants, oracles are still robust.}
  {\vspace{-0.2cm}}
  \label{tab:models_metrics_id}
  
  \resizebox{.9\linewidth}{!}{
\begin{tabular}{ll|ccc}
\toprule
\multicolumn{2}{c|}{Method} &
  \multicolumn{3}{c}{ID wildfire risk rasters} \\
Conditioning &
   Encoder &
  MSE $\downarrow$ &
  SSIM $\uparrow$ &
  MAE $\downarrow$ \\
\midrule
\multirow{3}{*}{Image only} &
  AlphaEarth &
  {\ul 0.026} &
  0.544 &
  {\ul 0.113} \\
 &
  SegFormer &
  0.034 &
  {\ul \textbf{0.558}} &
  0.131 \\
 &
  U-Net &
  0.062 &
  0.498 &
  0.186 \\
  \midrule
\multirow{3}{*}{+ Climate Data} &
  AlphaEarth &
  {\ul \textbf{0.025}} &
  0.552 &
  {\ul \textbf{0.110}} \\
 &
  SegFormer &
  0.038 &
  0.533 &
  0.131 \\
 &
  U-Net &
  0.042 &
  0.518 &
  0.144 \\
  \midrule
\multirow{3}{*}{+ Oracle} &
  AlphaEarth &
  0.043 &
  0.534 &
  0.150 \\
 &
  SegFormer &
  0.044 &
  0.538 &
  0.152 \\
 &
  U-Net &
  0.050 &
  0.507 &
  0.169 \\
  \midrule
\multirow{3}{*}{+ CoT}
 &
  AlphaEarth &
  {\ul 0.028} &
  {\ul 0.547} &
  {\ul 0.119} \\
 &
  SegFormer &
  0.039 &
  {\ul  0.548} &
  0.140 \\
 &
  U-Net &
  0.042 &
  0.534 &
  0.152 \\
  \bottomrule
\end{tabular}
}


\vspace{-10pt}
\end{table}

\subsection{Performance Results}\label{subsec:Results}

\noindent\textbf{Out-Of-Distribution (OOD) Raster Results.} A comparison between Encoder-Decoders with different conditionings out-of-distribution is provided in Table~\ref{tab:models_metrics_ood}. This is where the FireScope framework truly shines: FireScope's CoT Oracles exhibit striking gains in OOD settings, while remaining robust ID. A visualization of FireScope's predicted risk across Europe is provided in Figure~\ref{fig:europe_risk_map}.


When distinguishing between wildfire \emph{events}, conditioning on CoT Oracle yields the best Brier and ROC AUC scores across \emph{every} vision backbone. Conditioning on Oracles without CoT generally also yields OOD gains, though not as good as with CoT. Interestingly, U-Net without conditioning yields the best ECE score -- apart from that, the addition of Oracle and CoT generally improves ECE as well.

Distinguishing between individual wildfire \emph{pixels}, the addition of Oracle once again leads to systematic improvement, with CoT Oracle acquiring the best ROC AUC and IoU scores. This is surprising -- although Oracle conditions on a single scalar, it is able to improve generated rasters on a pixel-level scale.

\textbf{Oracle Results.} A comparison between Oracles is provided in Table~\ref{tab:oracle_metrics}. Looking at the ID performance of Oracles, we observe something striking: the Climate MLP Oracle achieves a QWK score of 0.76, comparable only to CoT Qwen, corresponding to substantial agreement with the ground truth~\cite{cohen1968weightedkappa}. Unlike for CoT Qwen, its performance does not transfer to OOD data. 
CoT Qwen outperforms its non-CoT counterpart, providing further evidence that explicit reasoning helps wildfire risk prediction. The rest of the Oracles perform significantly worse.

\noindent\textbf{In-Distribution (ID) Raster Results.} A comparison between Encoder-Decoders with different conditionings in-distribution is provided in Table~\ref{tab:models_metrics_id}. Across all Encoder-Decoder versions, those conditioned on raw climate vectors tend to slightly outperform the rest. 
Conditioning on CoT Oracles is generally on par with other baselines across metrics, indicating neither systematic gains nor significant losses in ID performance incurred by the FireScope framework. Oracles without CoT perform slightly worse, in line with the difference in Oracle QWK scores. 

\subsection{Ablation Study} \label{subsec:ablation_study}

\begin{figure}[ht!]
\vspace{-0.5cm}
    \centering
    \includegraphics[width=\linewidth]{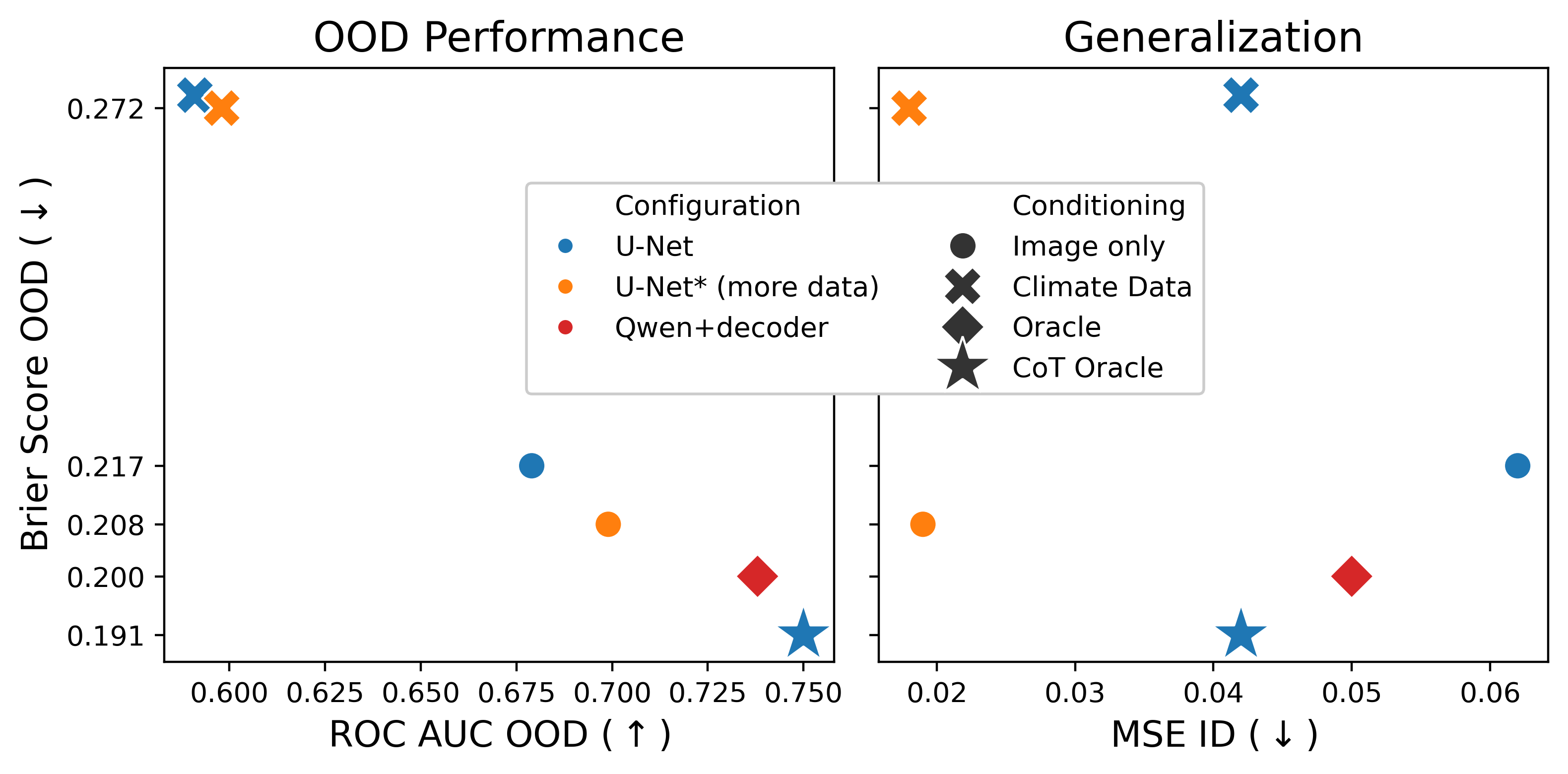}\vspace{-0.8cm}
    \setlength{\belowcaptionskip}{-10pt}
    \caption{Ablation study. We assess the effects of more training data on our U-Net model (with and without climate conditioning). We also attempt to attach a decoder head to Qwen directly without conditioning a separate Encoder-Decoder. FireScope remains the optimal configuration.}
    \label{fig:ablation}
\end{figure}

We report the result of different ablations in Figure~\ref{fig:ablation}. 

Firstly, we assess the effect of data size, training our U-Net on the whole volume of training data with translational augmentations (see section \ref{sec:benchmark}) -- 40 times larger in total. While it exhibits improved ID performance, it remains inferior to FireScope's CoT Oracle-conditioned configuration OOD. Secondly, we evaluate whether the VLM, freed from the scalar bottleneck, could independently learn to generate risk masks surpassing those of FireScope. For that, we directly apply a perceiver-style decoder~\cite{jaegle2022perceiveriogeneralarchitecture} to the last embedding layer of Qwen and train it supervisedly on its generated rasters. We call it \emph{Qwen+decoder}. FireScope renders as the optimal configuration.

\subsection{Interpretability Experiments}

We conduct two experiments to quantify the interpretability and quality of reasoning traces generated by FireScope. Both experiments are conducted on CoT Oracle, which we refer to as simply Oracle in this section.

\begin{figure*}[t]
\setlength{\belowcaptionskip}{-10pt}
    \centering
    \includegraphics[width=\linewidth]{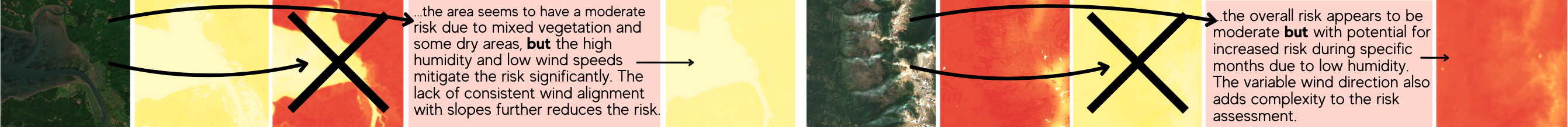}
    \captionof{figure}{Examples of failure cases when conditioning AlphaEarth on Oracle, fixed with the addition of CoT. Enabling iterative reasoning helps the model correct spurious correlations and align predictions with underlying physical and climatic drivers.}
    \label{fig:cots}
\end{figure*}

\subsubsection{Expert Study.} \label{subsubsec:expert_study}
To determine whether Oracle's CoTs contain meaningful cues, we design the following experiment. We sample 50 areas from the test set following the sampling procedure described in Section~\ref{sec:benchmark}. We then generate two CoTs for each area: the first is a default reasoning trace; the second, which we refer to as the \emph{golden} CoT, we generate by querying GPT-5 \emph{after giving it the correct risk classification ad-hoc}. Thus, we ask it to solve a much simpler task: knowing the correct risk classification, simulate the reasoning necessary to reach it. We then preprocess the CoTs by running them through Qwen/Qwen3-4B-Instruct-2507, prompting it to provide concise summaries and omit conclusions. This converts each CoT into a wildfire risk interpretation that highlights the factors considered, without alluding to a final classification. Finally, we shuffle and anonymize the interpretations, and reach out to two wildfire experts from the Wildfire Risk to Communities project~\cite{wrc}, asking them to provide a wildfire risk classification twice for each of the 50 areas: once based on Oracle's interpretation, and once based on the golden interpretation. We then measure the experts' QWK scores. This allows us to assess how much signal the interpretations carry, while also providing a golden reference for comparison.

\subsubsection{Automated Metrics.} \label{subsubsec:automated_metrics}
While there exist established metrics to definitively evaluate model performance, examining the level of interpretability in a model's CoT remains largely an unsolved problem. However, recent work has introduced methods that, albeit not conclusive, shed light on the faithfulness of models to their CoT~\cite{lanham2023measuring}. Inspired by this work, we document two interpretability metrics, both measured by synthetically modifying Oracle's CoTs and measuring the change in its final risk classification.

\textbf{Fidelity.} To quantify how much Oracle is actively guided by its CoT, we take the reasoning traces and risk rasters produced by U-Net FireScope on the testing set. We prompt Qwen/Qwen3-4B-Instruct-2507 to perturb each trace so that it argues for the risk level farthest from the prediction without changing facts. We then inject the perturbed CoTs into Oracle, acquire new classifications, and generate new risk rasters with U-Net FireScope. We measure how much the prediction shifts \emph{toward the risk level that the perturbed CoT alludes to}. We define:
\begin{equation}
\mathrm{fid} = \frac{1}{N} \sum_{i=1}^N \frac{(\tilde{y_i} - y_i)}{({y_i}^\ast - y_i)} \in [-1, 1]
\end{equation}
where $N$ is number of pixels, $y_i$ is the original prediction at pixel $i$, $\tilde{y_i}$ is a perturbed prediction, ${y_i}^\ast = 1.0$ if $y_i < 0.5$ and ${y_i}^\ast = 0$ otherwise. We refer to this metric \textit{fidelity}; it quantifies how much Oracle is actively guided by its CoT.

\textbf{Consistency.} To assess whether FireScope is guided by its CoT in a human-interpretable way, we prompt Qwen/Qwen3-4B-Instruct-2507 to rewrite each CoT in the testing set with different phrasing while preserving its factual content and logical meaning. We then inject these paraphrased CoTs and obtain new risk rasters. We compute the magnitude of the shift between original and paraphrased predictions. The consistency score is defined as:
\begin{equation}
\mathrm{cons} = 1-\frac{1}{N} \sum_{i=1}^N \frac{|\hat{y_i} - y_i|}{d_i}\in [0, 1], 
\end{equation}
where $N$ is number of pixels, $y_i$ denotes the original prediction at pixel $i$, $\hat{y_i}$ is the prediction with the paraphrased CoT, and $d_i=y_i$ if $\hat{y_i} < y_i$, $d_i=1-y_i$ if $\hat{y_i} > y_i$, and $d_i=1$ if $\hat{y_i} = y_i$. Higher values of consistency indicate that if the model relies on its CoT, it does so in a human-interpretable way.

\begin{table}[t]
\centering
\caption{Results of interpretability evaluations. ``Golden'' provides reference values of expert study. Experts are able to recover a significant amount of the risk level based solely on factors considered by Oracle. FireScope also has near-perfect consistency and high fidelity, indicating that raster generation is interpretably guided by Oracle's CoT.}
{\vspace{-0.3cm}}
\resizebox{\columnwidth}{!}{
\begin{tabular}{l|cc|cc}
\toprule
\multirow{2}{*}{Source} &
\multicolumn{2}{c|}{Expert Study} &
\multicolumn{2}{c}{Automated Metrics} \\
 & Exp.~1~QWK $\uparrow$& Exp.~2~QWK $\uparrow$& Fidelity $\uparrow$& Consistency $\uparrow$\\
\midrule
Oracle & 0.33 & 0.11 & 0.33 & 0.91 \\
Golden & 0.50 & 0.59 & n/a & n/a \\
\bottomrule
\end{tabular}}
\label{tab:interpretability_results}
\vspace{-0.3cm}
\end{table}

\subsubsection{Interpretability Results}
The results of all interpretability experiments are presented in Table~\ref{tab:interpretability_results}. While Oracle's CoT does not carry as much signal as the golden reasoning traces, the risk factors considered by it allowed experts to partially recover the risk scores. One expert achieved 0.33 QWK -- nearly 70\% of their score on the golden traces. The other expert achieved 0.11, which, albeit significantly less, is still above zero, indicating statistical significance. Further, FireScope achieves near-perfect consistency, meaning that injecting paraphrased CoTs barely affects the generated raster. Its fidelity is 0.33 -- the pixel-wise predicted risk shifts by 33\% toward the opposite scale on average when Oracle's CoT is perturbed. Because there is only so much the CoT can be changed without affecting factual information of the area's features, it is expected that fidelity cannot be as good as consistency. Additionally, we illustrate the effect of CoT in Figure~\ref{fig:cots}. Together, these results indicate that the raster generation process is actively guided by Oracle's CoT in a human-interpretable way. 
\section{Discussion}
\label{sec:discussion}

The results in Section~\ref{sec:experiments} reveal several noteworthy insights. The strong in-distribution performance of climate-conditioned models underscores a potential overfitting issue: the coarse spatial resolution of the climatology data allows models to memorize regional climate signatures rather than learning generalizable physical relationships. This makes FireScope-Bench interesting, as it naturally probes the tension between climatic overfitting and true generalization.

The performance of the FireScope method indicates that explicit reasoning enhances robustness without sacrificing in-domain accuracy. Notably, FireScope improves not only coarse regional classification but also per-pixel generalization: the Encoder-Decoder learns to leverage the Oracle’s reasoning as a contextual prior, rather than simply concatenating Oracle outputs as auxiliary metadata. The OOD benefits persist even when compared to models trained on forty times more data, demonstrating that structured reasoning contributes generalization beyond what data scaling alone can achieve. Finally, the inferior performance of the Qwen+decoder ablation -- where a decoder head is attached directly to the VLM -- indicates that, despite contributing only a scalar output, explicit reasoning provides a more structured and semantically grounded form of conditioning. This suggests that FireScope’s gains do not stem from the raw representational capacity of the VLM.

The expert study and automated evaluations show that FireScope's CoTs encode meaningful, human-understandable cues -- allowing domain experts to partially reconstruct risk judgments. Notably, there seems to be a level of subjectivity to the usefulness of FireScope's CoTs, as they resulted in varying expert performance, unlike the golden CoTs. High consistency and non-trivial fidelity further demonstrate that the model responds coherently to perturbations of its CoT, providing evidence that FireScope’s outputs remain both faithful to and interpretable through the Oracle’s reasoning.

\textbf{Limitations.} While FireScope demonstrates strong gains in robustness and interpretability, the communication between the Encoder-Decoder and the Oracle is bottlenecked by a single scalar signal. While this makes the system easy to interpret, the Oracle can only convey limited spatial granularity, constraining the extent to which fine-scale patterns can be informed by its CoT reasoning. Exploring architectures that allow the Oracle to provide CoT-enhanced, spatially resolved, or multi-dimensional conditioning -- for instance, through token-level or region-aware embeddings -- represents an exciting direction for future work toward more expressive reasoning-guided geospatial modeling.


\section{Conclusion}
\label{sec:conclusion}
We present \emph{FireScope} and \emph{FireScope-Bench}, introducing reasoning-driven wildfire risk prediction as a new frontier in multimodal vision research. Through comprehensive evaluations, we demonstrate that coupling a vision–language Oracle with chain-of-thought (CoT) reasoning to a fine-grained vision Encoder–Decoder yields consistent in-distribution performance and substantial out-of-distribution gains. FireScope improves even per-pixel generalization, indicating that the Oracle’s reasoning acts as a contextual prior -- guiding the model’s internal attention and enhancing spatial confidence rather than serving as a mere auxiliary cue. Remarkably, the OOD benefits persist even when compared to models trained on orders of magnitude more data, underscoring the efficiency and robustness of reasoning-based conditioning. Our ablations further confirm that the observed gains arise not from the inclusion of a VLM alone, but from the integration of interpretable reasoning within the FireScope framework.

Together, these results establish FireScope as both a testbed and methodology for studying how explicit reasoning can enhance generalization in geospatial prediction.

\section{Acknowledgements}
This research was partially funded by the Ministry of Education and Science of Bulgaria (support for INSAIT, part of the Bulgarian National Roadmap for Research Infrastructure).

We would like to thank the experts from the US Department of Agriculture’s Fire Modeling Institute for their valuable insights and contributions. In particular, we are deeply grateful to Jeffrey Chandler and Timothy Tallmadge for participating in the interpretability expert study and for their direct assistance in evaluating FireScope.

We would also like to express our deepest gratitude to Radena Gaydarova, whose artistic eye helped shape the visuals in this work.

{
    \small
    \bibliographystyle{ieeenat_fullname}
    \bibliography{main}

@String(CVPR= {IEEE Conf. Comput. Vis. Pattern Recog.})

@String(ICCV= {Int. Conf. Comput. Vis.})

@String(ECCV= {Eur. Conf. Comput. Vis.})

@String(ICLR = {Int. Conf. Learn. Represent.})

@String(AAAI = {AAAI})

@String(CVPR  = {CVPR})

@String(ICCV  = {ICCV})

@String(ECCV  = {ECCV})

@String(ICLR  = {ICLR})

@article{kondylatos2022wildfire,
  title={Wildfire danger prediction and understanding with deep learning},
  author={Kondylatos, Spyros and Prapas, Ioannis and Ronco, Michele and Papoutsis, Ioannis and Camps-Valls, Gustau and Piles, Mar{\'\i}a and Fern{\'a}ndez-Torres, Miguel-{\'A}ngel and Carvalhais, Nuno},
  journal={Geophysical Research Letters},
  volume={49},
  number={17},
  pages={e2022GL099368},
  year={2022},
  publisher={Wiley Online Library}
}

@article{zhao2025trade,
  title={Trade-offs in large reasoning models: An empirical analysis of deliberative and adaptive reasoning over foundational capabilities},
  author={Zhao, Weixiang and Sui, Xingyu and Guo, Jiahe and Hu, Yulin and Deng, Yang and Zhao, Yanyan and Qin, Bing and Che, Wanxiang and Chua, Tat-Seng and Liu, Ting},
  journal={preprint arXiv:2503.17979},
  year={2025}
}

@article{arlovic2025evaluation,
  title={Evaluation of synthetic data impact on fire segmentation models performance},
  author={Arlovic, Matej and Hrzic, Franko and Patel, Mitesh and Bednarz, Tomasz and Balen, Josip},
  journal={Scientific Reports},
  volume={15},
  number={1},
  pages={16759},
  year={2025}
}

@inproceedings{
beery2018recognition, 
title={Recognition in Terra Incognita}, 
author={Beery, Sara and Van Horn, Grant and Perona, Pietro}, 
booktitle={European Conference on Computer Vision (ECCV)}, 
year={2018}
}

@inproceedings{
koh2021wilds, 
title={WILDS: A Benchmark of in-the-Wild Distribution Shifts}, 
author={Koh, Pang Wei and Sagawa, Shiori and Marklund, Henrik and Xie, Sang Michael and Zhang, Marvin and Balsubramani, Akhil and Hu, Weihua and Yasunaga, Michihiro and Liang, Percy and Carmon, Yair and others}, 
booktitle={International Conference on Machine Learning (ICML)}, 
year={2021}
}

@article{wei2022chain,
  title={Chain-of-Thought Prompting Elicits Reasoning in Large Language Models},
  author={Wei, Jason and Wang, Xuezhi and Schuurmans, Dale and Bosma, Maarten and Ichter, Brian and Xia, Fei and Chi, Ed and Le, Quoc and Zhou, Denny},
  journal={preprint arXiv:2201.11903},
  year={2022}
}

@article{kojima2022zeroshot,
  title={Large Language Models are Zero-Shot Reasoners},
  author={Kojima, Takeshi and Gu, Shixiang Shane and Reid, Machel and Matsuo, Yutaka and Iwasawa, Yusuke},
  journal={preprint arXiv:2205.11916},
  year={2022}
}

@article{wang2022selfconsistency,
  title={Self-Consistency Improves Chain-of-Thought Reasoning in Language Models},
  author={Wang, Xuezhi and Wei, Jason and Schuurmans, Dale and Bosma, Maarten and Chi, Ed and Le, Quoc and Zhou, Denny},
  journal={preprint arXiv:2203.11171},
  year={2022}
}

@inproceedings{suris2023viper,
  title={{ViperGPT}: Visual Inference via Python Execution for Reasoning},
  author={Sur{\'\i}s, D{\'a}vid and Menon, Sachit and Vondrick, Carl},
  booktitle={IEEE/CVF International Conference on Computer Vision (ICCV)},
  year={2023}
}

@article{liu2023mmreact,
  title={{MM-ReAct}: Prompting {ChatGPT} for Multimodal Reasoning and Action},
  author={Liu, Haotian and Li, Chunyuan and Zhang, Pengchuan and Lee, Yong Jae},
  journal={preprint arXiv:2303.11381},
  year={2023}
}

@article{zhang2023visualcot,
  title={Visual Chain-of-Thought Reasoning for Multimodal Large Language Models},
  author={Zhang, Renrui and Li, Zheng and Li, Hongyang and Qiao, Yu and Gao, Peng},
  journal={preprint arXiv:2309.17421},
  year={2023}
}

@article{shao2024deepseekmath,
  title        = {{DeepSeekMath}: Pushing the Limits of Mathematical Reasoning in Open Language Models},
  author       = {Shao, Zhihang and Wang, Ziyu and Zhang, Yuxin and Zheng, Zihan and Liu, Yao and Liu, Zihan and Shang, Yibo and Xu, Linyang and Zhang, Tianyang and Chen, Lingpeng and others},
  journal      = {preprint arXiv:2402.03300},
  year         = {2024}
}

@article{jin2024reasonpix2pix,
  title={Instruction Reasoning Dataset for Advanced Image Editing},
  author={Jin, Ying and Ling, Pengyang and Dong, Xiaoyi and Zhang, Pan and Wang, Jiaqi and Lin, Dahua},
  journal={preprint arXiv:2405.11190},
  year={2024}
}

@article{lin2024vqascore,
  title={Evaluating Text-to-Visual Generation with Image-to-Text Models},
  author={Lin, Ziqiu and others},
  journal={preprint arXiv:2404.01291},
  year={2024}
}

@article{meng2024phybench,
  title={{PhyBench}: A Physical Commonsense Benchmark for Evaluating Text-to-Image Models},
  author={Fanqing Meng and Wenqi Shao and Lixin Luo and Yahong Wang and Yiran Chen and Quanfeng Lu and Yue Yang and Tianshuo Yang and Kaipeng Zhang and Yu Qiao and Ping Luo},
  journal={preprint arXiv:2406.11802},
  year={2024}
}

@article{kajic2024numerical,
  title={Evaluating Numerical Reasoning in Text-to-Image Models},
  author={Kaji{\'c}, Ivan and others},
  journal={preprint arXiv:2406.14774},
  year={2024}
}

@article{r2i2025,
  title={{R2I-Bench}: Benchmarking Reasoning-Driven Text-to-Image Generation},
  author={Chen, Kaijie and Lin, Zihao and Xu, Zhiyang and Shen, Ying and Yao, Yuguang and Rimchala, Joy and Zhang, Jiaxin and Huang, Lifu},
  journal={preprint arXiv:2505.23493},
  year={2025}
}

@inproceedings{ronneberger2015unet,
  title={{U-Net}: Convolutional Networks for Biomedical Image Segmentation},
  author={Ronneberger, Olaf and Fischer, Philipp and Brox, Thomas},
  booktitle={Medical Image Computing and Computer-Assisted Intervention (MICCAI)},
  year={2015}
}

@inproceedings{chen2018deeplab,
  title={Encoder-Decoder with Atrous Separable Convolution for Semantic Image Segmentation},
  author={Chen, Liang-Chieh and Zhu, Yukun and Papandreou, George and Schroff, Florian and Adam, Hartwig},
  booktitle={European Conference on Computer Vision (ECCV)},
  year={2018}
}

@inproceedings{perez2018film,
  title={{FiLM}: Visual Reasoning with a General Conditioning Layer},
  author={Perez, Ethan and Strub, Florian and de Vries, Harm and Dumoulin, Vincent and Courville, Aaron},
  booktitle={AAAI Conference on Artificial Intelligence},
  year={2018}
}

@inproceedings{esser2021vqgan,
  title={Taming Transformers for High-Resolution Image Synthesis},
  author={Esser, Patrick and Rombach, Robin and Ommer, Björn},
  booktitle={IEEE/CVF Conference on Computer Vision and Pattern Recognition (CVPR)},
  year={2021}
}

@inproceedings{xie2021segformer,
  title={{SegFormer}: Simple and Efficient Design for Semantic Segmentation with Transformers},
  author={Xie, Enze and Wang, Wenhai and Yu, Zhiding and Anandkumar, Anima and Alvarez, Jose M. and Luo, Ping},
  booktitle={Advances in Neural Information Processing Systems (NeurIPS)},
  year={2021}
}

@inproceedings{rombach2022ldm,
  title={High-Resolution Image Synthesis with Latent Diffusion Models},
  author={Rombach, Robin and Blattmann, Andreas and Lorenz, Dominik and Esser, Patrick and Ommer, Björn},
  booktitle={IEEE/CVF Conference on Computer Vision and Pattern Recognition (CVPR)},
  year={2022}
}

@inproceedings{balaji2022ediffi,
  title={{eDiff-I}: Text-to-Image Diffusion Models with an Ensemble of Expert Denoisers},
  author={Balaji, Yogesh and et al.},
  booktitle={preprint arXiv:2211.01324},
  year={2022}
}

@techreport{VanWagner1987_FWI,
  author      = {Van Wagner, C. E.},
  title       = {Development and Structure of the Canadian Forest Fire Weather Index System},
  institution = {Canadian Forestry Service, Petawawa National Forestry Institute},
  address     = {Chalk River, Ontario},
  year        = {1987},
  number      = {Forestry Technical Report 35},
  pages       = {37},
  url         = {https://cfs.nrcan.gc.ca/pubwarehouse/pdfs/19927.pdf}
}

@article{Bohlmann2024NGRFWI,
  title        = {Statistical calibration of probabilistic medium-range Fire Weather Index forecasts in Europe},
  author       = {Stephanie Bohlmann and Marko Laine},
  journal      = {Natural Hazards and Earth System Sciences},
  year         = {2024},
  volume       = {24},
  pages        = {4225--4235}
}

@article{TorresVazquez2025HybridSeasonal,
  title        = {Enhancing seasonal fire predictions with hybrid dynamical and random forest models},
  author       = {Miguel Ángel Torres-Vázquez and Sixto Herrera and Andrina Gincheva and Amar Halifa-Marín and Leone Cavicchia and Francesca Di Giuseppe and Juan Pedro Montávez and Marco Turco},
  journal      = {Natural Hazards},
  year         = {2025},
  volume       = {2},
  articlenumber= {20}
}

@article{McNorton2024PoF,
  title        = {A Global Probability-Of-Fire ({PoF}) Forecast},
  author       = {Joe Ramu McNorton and Francesca {Di Giuseppe} and Ewan Mark Pinnington and Matthew Chantry and Chris Barnard},
  journal      = {Geophysical Research Letters},
  year         = {2024},
  volume       = {51},
  pages        = {e2023GL107929}
}

@article{DiGiuseppe2025NatComms,
  title   = {Global data-driven prediction of fire activity},
  author  = {Di Giuseppe, Francesca and McNorton, Joe and Lombardi, Anna and Wetterhall, Fredrik},
  journal = {Nature Communications},
  year    = {2025},
  volume  = {16},
  number  = {1},
  pages   = {58097}
}

@article{UncAwareDL2025,
  title        = {Uncertainty-Aware Deep Learning for Wildfire Danger Forecasting},
  author       = {Spyros Kondylatos and Gustau {Camps-Valls} and Ioannis Papoutsis},
  year         = {2025},
  journal = {preprint arXiv:2509.25017}
}

@article{Shadrin2024SpreadDL,
  title        = {Wildfire spreading prediction using multimodal data and deep neural network approach},
  author       = {Dmitrii Shadrin and Svetlana Illarionova and Fedor Gubanov and Ksenia Evteeva and Maksim Mironenko and Ivan Levchunets and Roman Belousov and Evgeny Burnaev},
  journal      = {Scientific Reports},
  year         = {2024},
  volume       = {14},
  pages        = {2606}
}

@article{Buch2023SMLFire,
  title   = {{SMLFire1.0}: a stochastic machine learning model for fire frequency and size distributions across the Western United States},
  author  = {Buch, Jeremy and Fischer, Erich and Pe{\~n}a, Jorge and others},
  journal = {Geoscientific Model Development},
  year    = {2023},
  volume  = {16},
  pages   = {3407--3432}
}

@article{Liu2024SHAPXAI,
  title   = {Application of Remote Sensing and Explainable Artificial Intelligence for Wildfire Risk Zoning in the Mountainous Region of {Southwest} {China}},
  author  = {Jia Liu and Yukuan Wang and Yafeng Lu and Pengguo Zhao and Shunjiu Wang and Yu Sun and Yu Luo},
  journal = {Remote Sensing},
  year    = {2024},
  volume  = {16},
  number  = {19},
  pages   = {3602}
}

@article{Xu2025ISPRS,
  title   = {Deep learning for wildfire risk prediction: Integrating remote sensing and environmental data},
  author  = {Zhengsen Xu and Jonathan Li and Sibo Cheng and Xue Rui and Yu Zhao and Hongjie Heand  Haiyan Guan and Aryan Sharma and Matthew Erxleben and Ryan Chang and Linlin Xu},
  journal = {ISPRS Journal of Photogrammetry and Remote Sensing},
  year    = {2025},
  note    = {Early access}
}

@article{Sengupta2025XAI,
  title   = {Recent advances in explainable machine learning models for wildfires: From forecasting to burned area estimation},
  author  = {Sengupta, A. and others},
  journal = {Environmental Data Science},
  year    = {2025},
  note    = {In press}
}

@article{schulman2017ppo,
  title={Proximal Policy Optimization Algorithms},
  author={Schulman, John and Wolski, Filip and Dhariwal, Prafulla and Radford, Alec and Klimov, Oleg},
  journal={preprint arXiv:1707.06347},
  year={2017}
}

@article{haarnoja2019soft,
  title={Soft Actor-Critic Algorithms and Applications},
  author={Haarnoja, Tuomas and Zhou, Aurick and Hartikainen, Kristian and Tucker, George and Ha, Sehoon and Tan, Jie and Kumar, Vikash and Zhu, Henry and Gupta, Abhishek and Abbeel, Pieter and Levine, Sergey},
  journal={preprint arXiv:1812.05905},
  year={2019}
}

@article{guo2025deepseekr1,
  title={{DeepSeek-R1}: Incentivizing Reasoning Capability in {LLMs} via Reinforcement Learning},
  author={DeepSeek-AI},
  journal={preprint arXiv:2501.12948},
  year={2025}
}

@article{zhang2025towards,
  title        = {Towards Omnidirectional Reasoning with 360-R1: A Dataset, Benchmark, and {GRPO}-based Method},
  author       = {Zhang, Xinshen and Ye, Zhen and Zheng, Xu},
  journal      = {preprint arXiv:2505.14197},
  year         = {2025}
}

@article{bastani2023satlaspretrainlargescaledatasetremote,
      title={{SatlasPretrain}: A Large-Scale Dataset for Remote Sensing Image Understanding}, 
      author={Favyen Bastani and Piper Wolters and Ritwik Gupta and Joe Ferdinando and Aniruddha Kembhavi},
      year={2023},
      journal={preprint arXiv:2211.15660}
}

@article{brown2025alphaearthfoundationsembeddingfield,
      title={{AlphaEarth} Foundations: An embedding field model for accurate and efficient global mapping from sparse label data}, 
      author={Christopher F. Brown and Michal R. Kazmierski and Valerie J. Pasquarella and William J. Rucklidge and Masha Samsikova and Chenhui Zhang and Evan Shelhamer and Estefania Lahera and Olivia Wiles and Simon Ilyushchenko and Noel Gorelick and Lihui Lydia Zhang and Sophia Alj and Emily Schechter and Sean Askay and Oliver Guinan and Rebecca Moore and Alexis Boukouvalas and Pushmeet Kohli},
      year={2025},
      journal={preprint arXiv:2507.22291} 
}

@misc{wrc,
    title = {Wildfire Risk to Communities. https://wildfirerisk.org},
    author = {USDA Forest Service},
    url = {https://wildfirerisk.org},
    note = {Accessed 24.10.2025}
}

@misc{power,
    title={Data obtained from National Aeronautics and Space Administration ({NASA}) {Langley} {Research} {Center's} Prediction Of Worldwide Energy Resources ({POWER}), {NASA} {Earth} {Science} {Division}},
    author={{NASA}},
    url = {https://power.larc.nasa.gov},
    note = {Accessed 24.10.2025}
}

@misc{powerdata,
    title={Data obtained from the {POWER} Project's Climatology},
    author={{NASA}},
    url = {https://power.larc.nasa.gov/api/temporal/climatology},
    note = {Accessed 24.10.2025}
}

@article{loshchilov2017decoupled,
  title         = {Decoupled Weight Decay Regularization},
  author        = {Loshchilov, Ilya and Hutter, Frank},
  journal       = {preprint arXiv:1711.05101},
  year          = {2017}
}

@inproceedings{loshchilov2017sgdr,
  title     = {{SGDR}: Stochastic Gradient Descent with Warm Restarts},
  author    = {Loshchilov, Ilya and Hutter, Frank},
  booktitle = {International Conference on Learning Representations (ICLR)},
  year      = {2017}
}

@article{bai2025qwen2_5vl,
  title        = {{Qwen2.5-VL} Technical Report},
  author       = {Bai, Shuai and Chen, Keqin and Liu, Xuejing and Wang, Jialin and Ge, Wenbin and Song, Sibo and Dang, Kai and Wang, Peng and Wang, Shijie and Tang, Jun and Zhong, Humen and Zhu, Yuanzhi and Yang, Mingkun and Li, Zhaohai and Wan, Jianqiang and Wang, Pengfei and Ding, Wei and Fu, Zheren and Xu, Yiheng and Ye, Jiabo and Zhang, Xi and Xie, Tianbao and Cheng, Zesen and Zhang, Hang and Yang, Zhibo and Xu, Haiyang and Lin, Junyang},
  journal      = {preprint arXiv:2502.13923},
  year         = {2025},
}

@misc{sentinel,
    author = {Copernicus},
    title = {Sentinel-2 was accessed on 24.10.2025 from https://registry.opendata.aws/sentinel-2},
    note = {Accessed 24.10.2025},
    url = {https://registry.opendata.aws/sentinel-2},
}

@misc{effispaper,
author = {San-Miguel-Ayanz, J. and Schulte, Ernst and Schmuck, Guido and Camia, Andrea and Strobl, Peter and Libertà, Giorgio and Giovando, Cristiano and Boca, Roberto and Sedano, Fernando and Kempeneers, Pieter and McInerney, Daniel and Withmore, Ceri and Oliveira, Sandra and Rodrigues, Marcos and Durrant, Tracy and Corti, Paolo and Oehler, Friderike and Vilar, Lara and Amatulli, Giuseppe},
year = {2012},
month = {03},
pages = {},
title = {Comprehensive Monitoring of Wildfires in Europe: The European Forest Fire Information System ({EFFIS})},
journal = {Approaches to Managing Disaster - Assessing Hazards, Emergencies and Disaster Impacts}
}

@misc{effis,
    author = {Copernicus},
    title = {{EFFIS} burnt areas (by {MODIS}) was accessed on 24.10.2025 from https://forest-fire.emergency.copernicus.eu},
    note = {Accessed 24.10.2025},
    url = {https://forest-fire.emergency.copernicus.eu},
}

@article{Marc-Andre2009,
author = {Parisien, Marc-André and Moritz, Max A.},
title = {Environmental controls on the distribution of wildfire at multiple spatial scales},
journal = {Ecological Monographs},
volume = {79},
number = {1},
pages = {127-154},
year = {2009}
}

@article{Wang2004_SSIM,
  author  = {Z. Wang and A. C. Bovik and H. R. Sheikh and E. P. Simoncelli},
  title   = {Image Quality Assessment: From Error Visibility to Structural Similarity},
  journal = {IEEE Transactions on Image Processing},
  volume  = {13},
  number  = {4},
  pages   = {600--612},
  year    = {2004}
}

@article{cohen1968weightedkappa,
  author  = {Cohen, Jacob},
  title   = {Weighted kappa: Nominal scale agreement with provision for scaled disagreement or partial credit},
  journal = {Psychological Bulletin},
  volume  = {70},
  number  = {4},
  pages   = {213--220},
  year    = {1968}
}

@article{Brier1950,
  author  = {G. W. Brier},
  title   = {Verification of Forecasts Expressed in Terms of Probability},
  journal = {Monthly Weather Review},
  volume  = {78},
  number  = {1},
  pages   = {1--3},
  year    = {1950}
}

@article{Fawcett2006_ROC,
  author  = {T. Fawcett},
  title   = {An Introduction to {ROC} Analysis},
  journal = {Pattern Recognition Letters},
  volume  = {27},
  number  = {8},
  pages   = {861--874},
  year    = {2006}
}

@article{Jaccard1912_IoU,
  author  = {P. Jaccard},
  title   = {The Distribution of the Flora in the Alpine Zone},
  journal = {New Phytologist},
  volume  = {11},
  number  = {2},
  pages   = {37--50},
  year    = {1912}
}

@inproceedings{Naeini2015_ECE,
  author    = {M. P. Naeini and G. F. Cooper and M. Hauskrecht},
  title     = {Obtaining Well Calibrated Probabilities Using Bayesian Binning},
  booktitle = {AAAI Conference on Artificial Intelligence},
  volume    = {29},
  number    = {1},
  year      = {2015}
}

@article{lanham2023measuring,
  title   = {Measuring Faithfulness in Chain-of-Thought Reasoning},
  author  = {Lanham, Tamera and Chen, Anna and Radhakrishnan, Ansh and Steiner, Benoit and Denison, Carson and Hernandez, Danny and Li, Dustin and Durmus, Esin and Hubinger, Evan and Kernion, Jackson and Luko{\v{s}}i{\=u}t{\.e}, Kamil{\.e} and Nguyen, Karina and Cheng, Newton and Joseph, Nicholas and Schiefer, Nicholas and Rausch, Oliver and Larson, Robin and McCandlish, Sam and Kundu, Sandipan and Kadavath, Saurav and Yang, Shannon and Henighan, Thomas and Maxwell, Timothy and Telleen-Lawton, Timothy and Hume, Tristan and Hatfield-Dodds, Zac and Kaplan, Jared and Brauner, Jan and Bowman, Samuel R. and Perez, Ethan},
  journal = {preprint arXiv:2307.13702},
  year    = {2023}
}

@online{openai2025gpt5,
  author = {OpenAI},
  title = {Introducing {GPT-5}},
  year = {2025},
  month = aug,
  url = {https://openai.com/index/introducing-gpt-5/},
  note = {Accessed: Nov. 12, 2025}
}

@article{Hoessly2025On,
  author    = {Hoessly, Linard},
  title     = {On Misconceptions about the Brier Score in Binary Prediction Models},
  journal   = {preprint arXiv:2504.04906v4},
  year      = {2025}
}

@inproceedings{long2015fully,
  title={Fully Convolutional Networks for Semantic Segmentation},
  author={Long, Jonathan and Shelhamer, Evan and Darrell, Trevor},
  booktitle={IEEE Conference on Computer Vision and Pattern Recognition (CVPR)},
  pages={3431--3440},
  year={2015}
}

@article{badrinarayanan2017segnet,
  title={{SegNet}: A Deep Convolutional Encoder-Decoder Architecture for Image Segmentation},
  author={Badrinarayanan, Vijay and Kendall, Alex and Cipolla, Roberto},
  journal={IEEE Transactions on Pattern Analysis and Machine Intelligence (TPAMI)},
  volume={39},
  number={12},
  pages={2481--2495},
  year={2017}
}

@article{jaegle2022perceiveriogeneralarchitecture,
      title={{Perceiver IO}: A General Architecture for Structured Inputs \& Outputs},
      author={Andrew Jaegle and Sebastian Borgeaud and Jean-Baptiste Alayrac and Carl Doersch and Catalin Ionescu and David Ding and Skanda Koppula and Daniel Zoran and Andrew Brock and Evan Shelhamer and Olivier Hénaff and Matthew M. Botvinick and Andrew Zisserman and Oriol Vinyals and Joāo Carreira},
      year={2022},
      journal={preprint arXiv:2107.14795}
}

@inproceedings{yue2024mmmu,
  title     = {MMMU: A Massive Multi‐Discipline Multimodal Understanding and Reasoning Benchmark for Expert AGI},
  author    = {Yue, Xiang and Ni, Yuansheng and Zheng, Tianyu and Zhang, Kai and Liu, Ruoqi and Zhang, Ge and Stevens, Samuel and Jiang, Dongfu and Ren, Weiming and Sun, Yuxuan and Wei, Cong and Yu, Botao and Yuan, Ruibin and Sun, Renliang and Yin, Ming and Zheng, Boyuan and Yang, Zhenzhu and Liu, Yibo and Huang, Wenhao and Sun, Huan and Su, Yu and Chen, Wenhu},
  booktitle = {Proceedings of the IEEE/CVF Conference on Computer Vision and Pattern Recognition (CVPR) 2024},
  year      = {2024},
  pages     = {9556--9567},
  doi       = {10.1109/CVPR52733.2024.00913},
    url={https://mmmu-benchmark.github.io/},
    note={accessed: November 2025}
}

@misc{mtbs,
  title        = {Monitoring Trends in Burn Severity (MTBS): Burned Area Boundaries Dataset},
  author       = {{USDA Forest Service} and {U.S. Geological Survey}},
  year         = {2024},
  howpublished = {\url{https://www.mtbs.gov}},
  note         = {USFS Geospatial Technology and Applications Center (GTAC), Salt Lake City, UT and
                  USGS Earth Resources Observation and Science (EROS) Center, Sioux Falls, SD}
}
}
\clearpage
\clearpage
\maketitlesupplementary



\begin{figure}[t]
    \centering
    \includegraphics[width=\columnwidth]{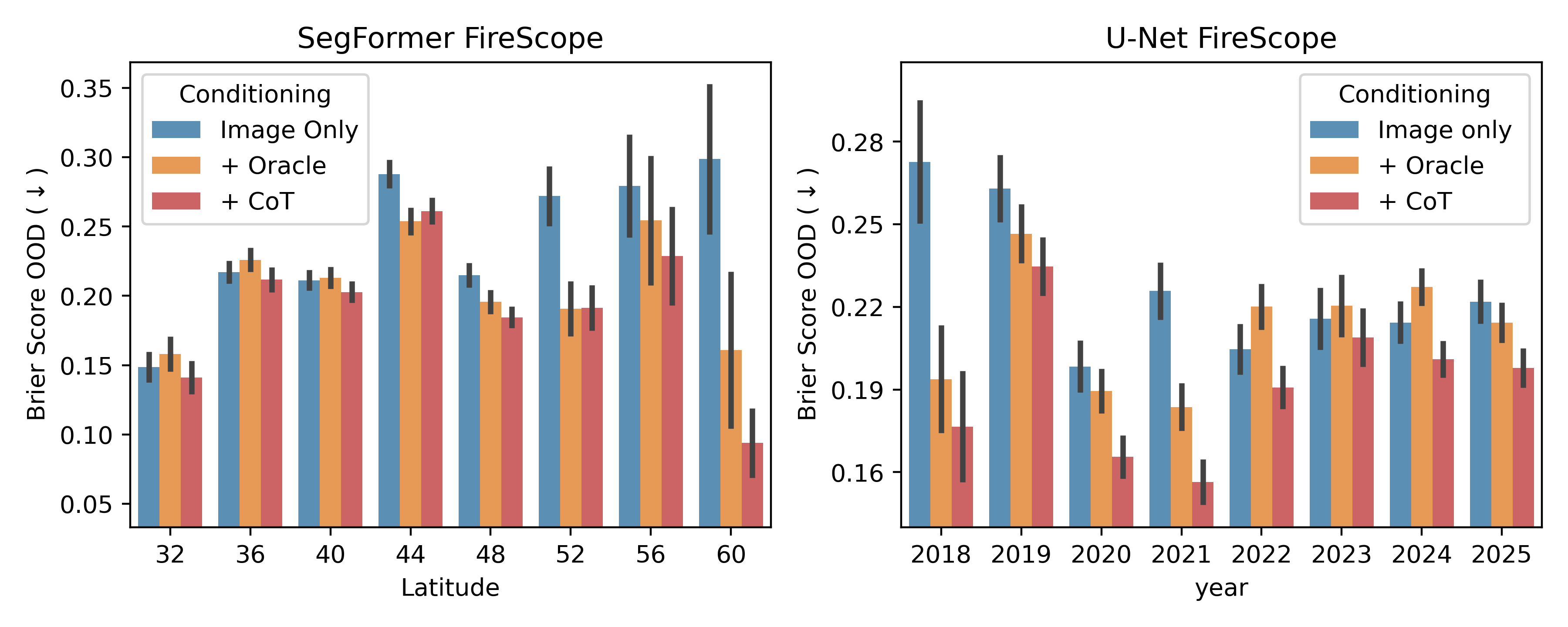}\vspace{-0.8cm}
    \setlength{\belowcaptionskip}{-15pt}
    \caption{Error distribution of FireScope in Europe across latitudes and years. FireScope's improvement is proportional to the distance from the distribution of training data, reaching its maximum in the 60th latitude and the year 2018. This suggests that FireScope is robust in varying operational settings.}
    \label{fig:errordist}
\end{figure}

\section{Wildfire events in the USA. Direct ID vs OOD comparison.}
For completeness and direct comparison betwen OOD and ID performance, we also include 2284 real wildfire events and 1522 control areas in the USA from MTBS~\cite{mtbs}, filtering out controlled fires. They are included in FireScope-Bench. Evaluation results are presented in Tab.~\ref{tab:direct_comparison}. This also allows for a direct comparison between ID and OOD performance. Comparing the difference in ROC AUC between FireScope and baseline, averaged across encoders, yields \textbf{-0.001 ID loss} against \textbf{+0.037 OOD gain} for wildfire events, and \textbf{-0.012 ID loss} against \textbf{+0.026 OOD gain} for pixels. Quantitatively, this demonstrates that FireScope's OOD gain significantly outweighs ID loss.
\begin{table}[hb]
\centering
  \caption{Model performance ID (MTBS fires in USA) and OOD (EFFIS fires in Europe). Top-3 are underscored. Top-1 is in bold.}
\vspace{-0.3cm}
  \resizebox{\linewidth}{!}{
\begin{tabular}{ll|cc|cc}
\toprule
\multicolumn{2}{c|}{Method} &
  \multicolumn{2}{c|}{OOD: ROC AUC $\uparrow$ } &
  \multicolumn{2}{c}{ID: ROC AUC $\uparrow$ } \\
Condition &
   Encoder &
  events &
  pixels &
  events &
  pixels \\
\midrule
\multirow{4}{*}{Image only} &
  AlphaEarth & 0.692 & 0.632 & \underline{\textbf{0.782}} & \underline{\textbf{0.633}} \\
 &
  SegFormer & 0.705 & 0.651 & 0.764 & \underline{0.613} \\
 &
  U-Net & 0.679 & 0.587 & 0.749 & 0.594 \\
  \midrule
\multirow{4}{*}{+ Climate} &
  AlphaEarth & 0.689 & 0.629 & \underline{0.778} & 0.573\\
 &
  SegFormer & 0.557 & 0.518 & 0.728 & 0.571 \\
 &
  U-Net & 0.591 & 0.559 & 0.750 & 0.575 \\
  \midrule
\multirow{4}{*}{+ Oracle} &
  AlphaEarth & 0.696 & \underline{0.653} & \underline{\textbf{0.782}} & 0.597 \\
 &
  SegFormer & 0.713 & 0.642 & 0.730 & 0.588\\
 &
  U-Net & 0.698 & \underline{0.655} & 0.743 & 0.610 \\
  \midrule

 &
  AlphaEarth & \underline{0.717} & \underline{0.653} & 0.766 & 0.596 \\
 + CoT  &
  SegFormer & \underline{0.727} & \underline{\textbf{0.658}} & 0.760 & 0.602 \\
 (FireScope) &
  U-Net & \underline{\textbf{0.750}} & 0.652 & 0.758 & \underline{0.615} \\
  \bottomrule
\end{tabular}
}
    \label{tab:direct_comparison}
\vspace{-0.6cm}
\end{table}

\section{Detailed Metrics}
\label{subsubsec:Detailed_Evaluation_Metrics}
\textbf{In-distribution (ID).} As we have ground truth continuous risk rasters in the US, we use three metrics for evaluation: 

\textit{Mean Squared Error (MSE)} to quantify per-pixel error:
\begin{equation}
    \text{MSE}=\frac{1}{N}\sum_i^{N}(x_i-y_i)^2
\end{equation}

\textit{Mean Absolute Error (MAE)} to quantify per-pixel prediction error:
\begin{equation}
    \text{MAE}=\frac{1}{N}\sum_i^{N}|x_i-y_i|
\end{equation}

\textit{Structural Similarity Index (SSIM)}~\cite{Wang2004_SSIM} to assess spatial and structural consistency between predicted and reference risk maps:
\begin{equation}
    \text{SSIM}=\frac{(2\mu_x\mu_y+C_1)(2\sigma_{xy}+C_2)}{(\mu_x^2+\mu_y^2+C_1)(\sigma_x^2+\sigma_y^2+C_2)}
\end{equation}
where $C_1=10^{-4}$, $C_2=9\cdot10^{-4}$, and $\mu_x$, $\sigma_x$, $\sigma_{xy}$ are local mean, standard deviation and covariance, calculated in a gaussian sliding window with size of 11 pixels.

\noindent\textbf{Out-of-Distribution (OOD).} Since we only have wildfire events in Europe, we use probabilistic and spatial metrics to assess OOD performance. Risk is a combination between the probability of burning and expected intensity, which is directly correlated to the probability of observing a wildfire in a satellite image -- places more likely to burn with greater intensity are consequently more likely to have observable fires. We therefore treat FireScope's predicted risk scores as probabilities of observable wildfire events.

\textit{Brier score} \cite{Brier1950} is equivalent to MSE on the binary labels $y$:
\begin{equation}
    \text{Brier Score}=\frac{1}{N}\sum_i^{N}(p_i-y_i)^2
\end{equation}

The \textit{Area Under the Receiver Operating Characteristic Curve (ROC AUC)} \cite{Fawcett2006_ROC} evaluates the model’s ability to discriminate between positive and negative samples:
\begin{equation}
    \text{ROC AUC}=\int^1_0\text{TPR}(\text{FPR}^{-1}(x))dx=P(X_1>X_0)
\end{equation}
where TPR is true-positive rate, FPR is false-positive rate, $X_1$ is score for positive labels and $X_0$ is background score.

\textit{Expected Calibration Error (ECE)} \cite{Naeini2015_ECE} quantifies the alignment between predicted probabilities and observed frequencies.
\begin{equation}
    \text{ECE}=\sum_{b\in \text{bins}}\frac{n_b}{N}|\text{acc}_b-\text{conf}_b|
\end{equation}
where $\text{bins}$ are 15 equally spaced bins over $[0,1]$, $\text{conf}_b$ is the mean of predicted probabilities inside the bin, and $\text{acc}_b$ is the accuracy of predictions whose predicted probabilities are inside the bin. $N$ is the total number of pixels, while $n_b$ is number of pixels inside bin $b$.

Finally, \textit{Intersection over Union (IoU)} \cite{Jaccard1912_IoU} measures the spatial overlap between predicted and reference burned areas:
\begin{equation}
    \text{IoU}=\frac{\text{TP}}{\text{TP}+\text{FP}+\text{FN}}
\end{equation}
where TP is true-positive, FP is false-positive, FN is false-negative.

Note that due to the stochastic nature of ignition, we treat non-burnt areas within wildfire tiles as background rather than negative samples~\cite{Marc-Andre2009}. Therefore, the maximum expected performance within individual images is limited, and predictions must be assessed based on the ability to distinguish wildfire events from controls areas in addition to quality of segmentation.

\noindent\textbf{Oracles.} To evaluate Oracles, we measure their classification performance on equivalent ordinal labels. We split risk level from $[0,1]$ into ten evenly spaced bins $\{0, 1, ..., 9\}$. For each training area, we take the corresponding ground truth risk raster and average the risk level across pixels. We thus map each area to an ordinal label from 0 to 9, representing the overall risk for that area. We then use the same metrics as above, but omit SSIM and IoU, as these are continuous predictions. We instead substitute them with \textit{Quadratic Weighted Kappa (QWK)}~\cite{cohen1968weightedkappa} for ID data. QWK is a performance metric specifically designed for ordinal classification tasks.

Formally, given $N$ items and $K$ ordinal labels, let $O$ be the observed $K \times K$ matrix of counts 
where $O_{ij}$ denotes the number of instances rated as class $i$ by the model 
and class $j$ in the ground truth. Let $E$ be the expected count matrix under the assumption of independent ratings, i.e.
\begin{equation}
E_{ij} = \frac{\big(\sum_{k} O_{ik}\big)\big(\sum_{k} O_{kj}\big)}{N}.
\end{equation}
The QWK score is then defined as
\begin{equation}
\kappa = 1 - \frac{\sum_{i,j} {(i-j)^2} O_{ij}}{\sum_{i,j} {(i-j)^2} E_{ij}}.
\end{equation}
Its values range from $-1$ (systematic disagreement) through $0$ (chance-level agreement) up to $1$ (perfect agreement). QWK measures \textbf{statistical agreement}, making it more sensitive to label distribution and chance-level predictions: for example, a naive model that always predicts the mean of the scale may achieve a deceptively high score in other metrics, but its QWK would remain close to zero, correctly reflecting the lack of real agreement with the ground truth.

\section{Experiments Configurations}
\subsection{Models}
\textbf{Oracles}. We select Qwen2.5-VL-7B-Instruct \cite{bai2025qwen2_5vl} 
as our Oracle for its strong open-source performance across multimodal reasoning benchmarks and efficiency to train and deploy at only 7B parameters. We train two versions of it: one with CoT reasoning as outlined in Section 4.1 of the main paper, and one trained with supervised fine-tuning without CoT. We evaluate the classification performance of three additional Oracles:
\begin{itemize}
  \item \emph{GPT-5}~\cite{openai2025gpt5}: a powerful general-purpose VLM.
  \item \emph{Fire Weather Index (FWI)}~\cite{VanWagner1987_FWI}: a well-established physics-based index that estimates fire danger based on meteorological variables. We average it across summers between 2016 and 2025.
  \item \emph{Climate MLP}: training a small multilayer perceptron head to classify based solely on the climatology vector.
\end{itemize}

\textbf{Encoders.} To evaluate the effectiveness of FireScope, we integrate the state-of-the-art vision encoder -- SegFormer~\cite{xie2021segformer}, offering complementary strengths in visual representation. We use SegFormer MiT-B5 version. To integrate Oracle’s output, we attach a FiLM-conditioned transformer decoder, which decodes hierarchically with cross-attention and modulation at multiple scales, using a learned target grid and skip fusions normalized with GroupNorm, followed by a small convolutional head that upsamples to the final resolution. We keep the pre-trained encoders frozen and fine-tune the FiLM and decoder head parameters.

Second, we include in our comparison AlphaEarth~\cite{brown2025alphaearthfoundationsembeddingfield}, a satellite embedding dataset in 10m resolution, as a strong in-domain baseline. The embedding model was pretrained on a number of tasks: self-distillation, natural language supervision, supervised reconstruction of land cover use, vegetation features, and monthly-aggregated climate. As it uses only Sentinel satellite imagery, it falls under the FireScope-Bench workflow. For training and evaluation, we concatenate AlphaEarth embeddings with the conditioning features and apply a simple two-layer convolutional neural network to produce the risk predictions.

Last, we design a lightweight U-Net~\cite{ronneberger2015unet} as a baseline trained from scratch. It uses four encoder and three decoder stages built from convolutional and residual blocks. Each encoder stage downsamples by a factor of 2-3 using strided convolutions, while feature depth doubles at each level. The decoder mirrors this hierarchy with transposed convolutions and skip connections that concatenate encoder activations at matching resolutions, followed by residual refinement blocks. The final 3×3 convolution produces the prediction map, cropped to match the target size. FiLM conditioning is applied to every stage.

\textbf{Conditionings.} For each choice of Encoder-Decoder, we train up to four versions:
\begin{itemize}
  \item Baseline: a baseline trained only on satellite images.
  \item Climate: conditioning on the raw climate vector.
  \item Oracle: conditioning on Qwen fine-tuned without CoT.
  \item CoT Oracle: conditioning on Qwen fine-tuned with CoT.
\end{itemize}

\textbf{Ablations.} We further include several additional experiments for ablation purposes:
\begin{itemize}
  \item Qwen-enc: adding a transformer decoder with a lightweight convolutional upsampler head directly to Qwen2.5-VL-7B-Instruct to generate the raster in one forward pass.
  \item U-Net*: training the U-Net on the full volume of training data with translational augmentations (see Section
  3 of the main paper).
\end{itemize}

We test each model ID on the 3935 testing set areas, and OOD on all wildfire and control events in Europe (see Section 3 of the main paper).

\subsection{CoT Oracle Fine-Tuning}\label{subsec:oracle_fine_tuning}
 Whilst the ordinal labels defined in Section 3.2 of the main paper
 can be used to perform supervised fine-tuning on the Oracle, this would restrict it to outputting a single scalar score. To allow it to explore intermediate reasoning steps and refine its estimate iteratively, we fine-tune it with reinforcement learning (RL), which is unhindered by the lack of gradients for arbitrary-length outputs. For that, we use group-relative policy optimisation (GRPO) \cite{shao2024deepseekmath} -- a novel reinforcement learning algorithm which foregoes the need for a critic model, thus incurring much less overhead compared to other widely used RL algorithms~\cite{schulman2017ppo, haarnoja2019soft}. 

In it, a reward function $R$ is defined that quantifies the performance of model outputs after they are complete. Given a policy model $\pi_{\theta_{\mathrm{old}}}(\cdot \mid p)$ parametrized by parameters ${\theta_{\mathrm{old}}}$ and sampled training input $p$, GRPO samples a group of outputs $o_1,...,o_n \sim\pi_{\theta_{\mathrm{old}}}(\cdot \mid p)$ and estimates advantages $\hat{A}_i$ by comparing individual rewards $r_i = R(o_i)$ to the group's normalized mean:
\setlength{\abovedisplayskip}{2pt}  
\setlength{\belowdisplayskip}{2pt}  
\begin{equation}\label{eq:grpo_advantage}
\hat{A}_i = \frac{r_i - b}{\sigma},\;
b = \tfrac{1}{n}\sum_{i=1}^n r_i,\;
\sigma = \sqrt{\tfrac{1}{n}\sum_{i=1}^n (r_i - b)^2}
\end{equation}
Then, the following objective is maximized:
\setlength{\abovedisplayskip}{2pt}  
\setlength{\belowdisplayskip}{2pt}  
\begin{equation}
\begin{alignedat}{1}
&\mathcal{J}_{\mathrm{GRPO}}(\theta)
= \mathbb{E}_{p \sim \mathcal{X},\, \{o_i\}_{i=1}^n \sim \pi_{{\theta_{\mathrm{old}}}}(\cdot \mid p)} \\[-0.25em]
&\quad \Big[
\frac{1}{n}\sum_{i=1}^n
\min\!(
d_i\,\hat{A}_i,\,
\mathrm{clip}(d_i,1-\varepsilon,1+\varepsilon)\,\hat{A}_i
) \\[-1em]
&\hspace{9.5em}\quad
- \beta\,D_{\mathrm{KL}}\!\big(\pi_{\theta}\|\pi_{\mathrm{ref}}\big)
\Big]
\label{eq:important}
\end{alignedat}
\end{equation}
where
\begin{equation}\label{eq:grpo_ratio}
d_i = \frac{\pi_{\theta}(o_i \mid p)}{\pi_{{\theta_{\mathrm{old}}}}(o_i \mid p)}
\end{equation}
\begin{equation}\label{eq:grpo_kl}
D_{\mathrm{KL}}\!\big(\pi_{\theta}\,\|\,\pi_{\mathrm{ref}}\big)
= \mathbb{E}_{o \sim \pi_{\theta}(\cdot \mid p)}
\!\left[
\log \frac{\pi_{\theta}(o \mid p)}{\pi_{\mathrm{ref}}(o \mid p)}
\right]
\end{equation}
Here, $\pi_{\mathrm{ref}}$ is a frozen copy of the model before fine-tuning. By maximizing this objective, we update the policy to increase the probability of outputs with positive advantages and reduce it for those with negative ones. $D_{\mathrm{KL}}$ is a regularizing term that penalizes updates that change the policy too radically.

\textbf{Reward Design.}
Our reward function consists of three components: an accuracy term, a formatting term, and a label-frequency-based reweighting factor. The final reward is defined as
\begin{equation}
R = w(\hat{y}) \left( 0.9\,R_{\mathrm{acc}} + 0.1\,R_{\mathrm{fmt}} \right),
\end{equation}
where $w(\hat{y})$ is a class-dependent weight determined by the frequency of the ground-truth label $\hat{y}$ in the training set.

Given a correctly parsed model prediction $y$ and the corresponding ground-truth label $\hat{y}$, we compute the absolute error
\begin{equation}
d = \lvert y - \hat{y} \rvert.
\end{equation}
The accuracy reward is defined as
\begin{equation}
R_{\mathrm{acc}} =
\begin{cases}
1.0, & \text{if } d = 0,\\
0.5, & \text{if } d = 1,\\
0.1, & \text{if } d = 2,\\
0,   & \text{otherwise.}
\end{cases}
\end{equation}
The formatting reward is
\begin{equation}
R_{\mathrm{fmt}} =
\begin{cases}
1, & \text{if the generated output is parseable,}\\
0, & \text{otherwise.}
\end{cases}
\end{equation}

Let $f(c)$ denote the frequency of class $c$ in the training set, and let
\begin{equation}
f_{\min} = \min_c f(c), \qquad f_{\max} = \max_c f(c).
\end{equation}
We assign each class $c$ a weight by linearly interpolating between higher weights for rarer classes and lower weights for more frequent classes:
\begin{equation}
w(c) = 1 + \left(\frac{f_{\max}}{f_{\min}} - 1\right)
- \left(\frac{f_{\max}}{f_{\min}} - 1\right)
\frac{f(c) - f_{\min}}{f_{\max} - f_{\min}}.
\end{equation}
Thus, rarer labels receive larger rewards, which compensates for class imbalance during training.

\subsection{Training Setup.} 
We perform full model fine-tuning on the CoT Oracle using the GRPO framework with a learning rate of $1\times10^{-5}$, four generations per prompt, and mixed-precision (bfloat16) training. We set per-device batch size to 16. The output length is capped at 1024 tokens, and optimization uses $D_{\mathrm{KL}}=0.01$. We train for 45 epochs over two days on four NVIDIA H200 GPUs.

We fine-tune the Oracle without CoT using cross-entropy loss on the next-token prediction of the target risk digit. Training runs with mixed-precision (bfloat16), using a batch size of 16, and a learning rate of $1\times10^{-5}$, and lasts less than 24 hours on four NVIDIA H200 GPUs.

For the VLM-encoder, we train Qwen2.5-VL-7B-Instruct with mixed-precision (bfloat16), a batch size of 16, and a learning rate of $1\times10^{-5}$, using the same training objective as for the Encoder-Decoder, as described in Section
4.2 of the main paper
.

All Encoder-Decoder architectures are optimized with the AdamW optimizer~\cite{loshchilov2017decoupled} ($\beta_1{=}0.9$, $\beta_2{=}0.95$, weight decay $=10^{-5}$) and a cosine annealing learning rate schedule~\cite{loshchilov2017sgdr}, applying gradient clipping at a norm of~1.0. We use a per-device batch size of 32. For each, we customize the learning rate and total training epochs:
\begin{itemize}
  \item SegFormer: lr = $1\times10^{-5}$, epochs = 500;
  \item U-Net: lr = $1\times10^{-3}$, epochs = 1000;
  \item AlphaEarth: lr = $1\times10^{-3}$, epochs = 1000.
\end{itemize}

To isolate the effects of our ablations, we keep those consistent for each Encoder-Decoder across experiments. Each training lasts less than 24 hours on four NVIDIA H200 GPUs.

After each training, we benchmark the checkpoint with the best evaluation performance.

We acknowledge the importance of gauging the sensitivity of the various hyperparameters, but due to the extensive number of experiments already required and the significant computational cost associated with each training run, a full hyperparameter sensitivity analysis is beyond our practical budget. Thus, we rely on configurations grounded in common practices in the field.

\section{Ablation results}

Metrics of ablation methods are reported in Table \ref{tab:abltable}
\begin{table}[ht]
\caption{Metrics of ablation methods.}\label{tab:abltable}{\vspace{-0.4cm}}
\resizebox{\columnwidth}{!}{
\begin{tabular}{ll|ccc|cc|ccc}
\toprule
\multicolumn{2}{c|}{Method} & \multicolumn{3}{c|}{OOD wildfire events} & \multicolumn{2}{c|}{OOD wildfire pixels} & \multicolumn{3}{c}{ID wildfire risk rasters} \\
Conditioning &
   Encoder &
  Brier $\downarrow$ &
  ROC AUC $\uparrow$ &
  ECE $\downarrow$ &
  ROC AUC $\uparrow$ &
  IoU@0.5 $\uparrow$ &
  MSE $\downarrow$ &
  SSIM $\uparrow$ &
  MAE $\downarrow$  \\
\midrule
n/a            & VLM-enc   & 0.200     & 0.738     & 0.069    & 0.647           & 0.176           & 0.050  & 0.468  & 0.182 \\
Image only     & Unet*     & 0.208     & 0.699     & 0.058    & 0.619           & 0.173           & 0.019  & 0.620  & 0.090 \\
Climate Data   & Unet*     & 0.272     & 0.598     & 0.171    & 0.573           & 0.153           & 0.018  & 0.622  & 0.081 \\
\bottomrule
\end{tabular}}
\end{table}

\section{General-Purpose Capabilities of CoT Oracle (FireScope)} \label{sec:general_purpose}
After prompting CoT Oracle to provide its reasoning and overall risk score, it may be of interest to users to acquire answers to follow-up questions or continue conversing. For this reason, we perform simple quantitative and qualitative analyses of CoT Oracle's general purpose capabilities. In this section, we refer to CoT Oracle as simply Oracle.

\textbf{Quantitative analysis.} To quantitatively assess how much of Oracle's general-purpose capabilities have been sacrificed to specialize in wildfire risk prediction, we evaluate it on the multiple-choice subset of the Massive Multi-discipline Multimodal Understanding and Reasoning Benchmark (MMMU)~\cite{yue2024mmmu}. Oracle achieves an accuracy of \textbf{51.12\%}. This is a drop of 7.48\% from Qwen2.5-VL 7B's performance before fine-tuning, which is not insignificant, but CoT Oracle still attains a better accuracy than many other popular VLMs of comparable size. 
This suggests that even after fine-tuning, Oracle remains surprisingly competent across diverse multimodal tasks, retaining much of its original reasoning ability.

\textbf{Qualitative analysis.} To better understand Oracle's abilities beyond producing scalar wildfire risk scores, we explore its behavior in extended, free-form interactions that mirror realistic user queries and decision-making scenarios. We provide an example conversation where the user asks follow-up questions about building a house in the area after attaining Oracle's wildfire risk estimate. The area is at $48.7392 ^\circ N, -113.4151 ^\circ W$, the Saint Mary lakes in Montana.

\begin{wrapfigure}{l}{0.18\textwidth}  
    \vspace{-10pt}                     
    \includegraphics[width=\linewidth]{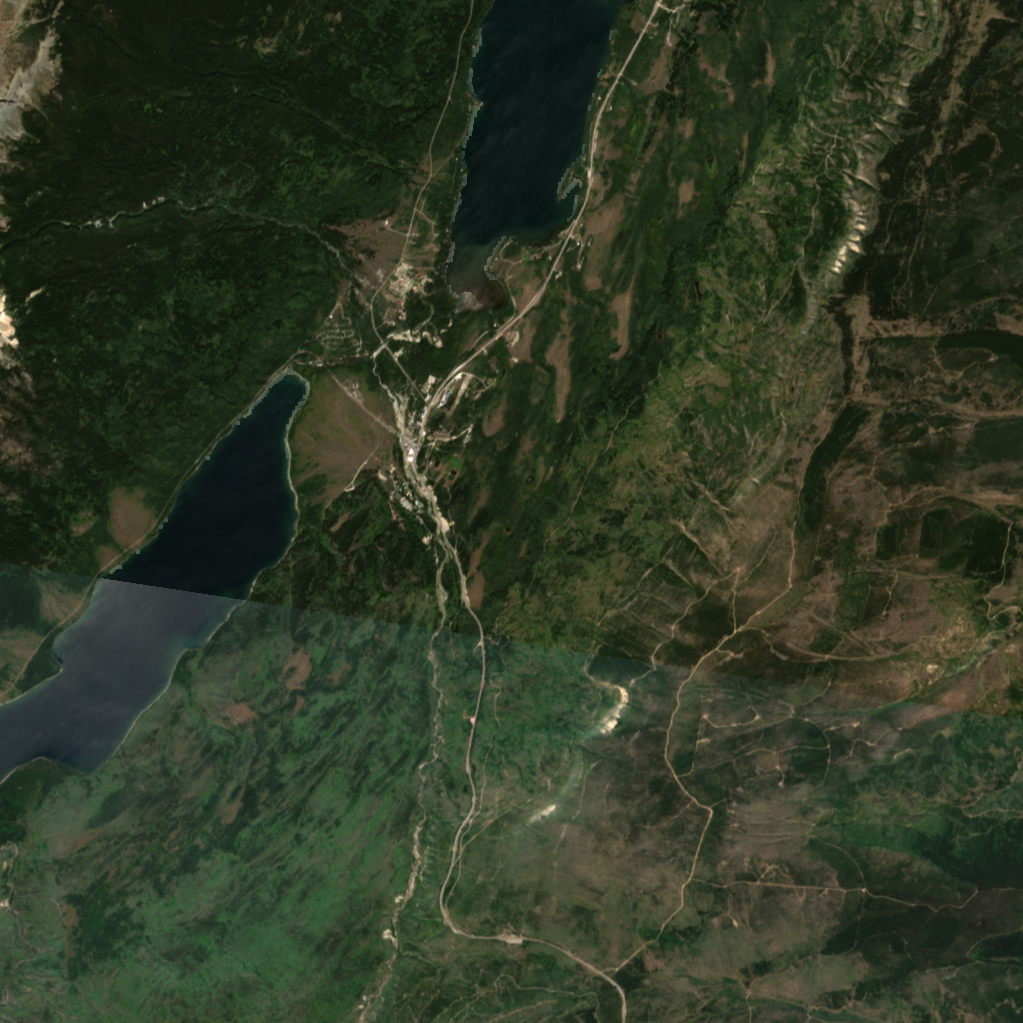}
    \vspace{-10pt}                     
\end{wrapfigure}

\footnotesize
\textbf{Example: Follow-up questions about the placement of a house in the area.}

\textbf{Oracle:} To determine the wildfire risk in the area based on the satellite image and climate data, we need to analyze the following factors:  \\
 \\
1. **Vegetation Density**: The satellite image shows areas with varying shades of green, indicating different levels of vegetation density. Dense vegetation can increase fire risk due to more fuel available for fires to spread. \\
 \\
2. **Climate Data**: \\
   - **Average Temperature**: Lower temperatures generally reduce fire risk because fires require higher temperatures to ignite and sustain. However, the average temperatures in the image region are relatively low, suggesting that fire risk might be moderate but not extremely high. \\
   - **Average Humidity**: Higher humidity reduces fire risk by providing moisture that can help extinguish fires. The climate data shows lower humidity in some months, which could increase fire risk. \\
   - **Average Wind Speed**: Stronger winds can increase fire risk by spreading flames faster. The climate data indicates variable wind speeds, with some months showing higher wind speeds, which could exacerbate fire spread. \\
   - **Average Wind Direction**: The wind direction is mostly from the west or northwest, which could align with potential slope directions in the area, increasing fire risk. \\
 \\
3. **Slope**: The satellite image shows areas with varying elevations, which could indicate slopes. Slopes can increase fire risk as they allow fires to spread more easily uphill. \\
 \\
Analysis: \\
 \\
- **Vegetation Density**: The image shows a mix of dense and less dense vegetation. Dense vegetation increases fire risk. \\
- **Climate Data**: \\
  - **Temperature**: Low temperatures suggest moderate fire risk. \\
  - **Humidity**: Lower humidity in some months suggests increased fire risk. \\
  - **Wind Speed**: Variable wind speeds, with some higher values, suggest increased fire risk. \\
  - **Wind Direction**: The wind direction is mostly from the west or northwest, which could align with potential slope directions, increasing fire risk. \\
- **Slope**: The image shows areas with varying elevations, which could indicate slopes. Slopes increase fire risk. \\
 \\
Risk Assessment: \\
 \\
- **Vegetation Density**: Moderate risk. \\
- **Climate Data**: \\
  - **Temperature**: Moderate risk. \\
  - **Humidity**: High risk. \\
  - **Wind Speed**: High risk. \\
  - **Wind Direction**: High risk. \\
- **Slope**: High risk. \\
 \\
Final Classification: \\
 \\
The overall risk assessment suggests a high level of wildfire risk in the area. The combination of dense vegetation, dry conditions, strong winds, and potential slopes all contribute to a high risk of wildfires. \\
 \\
FINAL ANSWER: \\
8 \\
 \\
\textbf{User:} If I had to build a house here, where would you recommend I build it and why? Give me a specific area in the image.  \\
 \\
\textbf{Oracle:} Given the high wildfire risk in the area, it's crucial to choose a location that minimizes exposure to potential fire hazards. Here are some recommendations based on the satellite image and the analysis of the wildfire risk factors: \\
 \\
1. **Avoid Areas Near Vegetation**: Stay away from areas with dense vegetation, as these are more likely to catch fire and spread quickly. \\
2. **Consider Elevation**: Higher elevation areas may have better air circulation, which can help dissipate heat and reduce the risk of fire. \\
3. **Avoid Low-Lying Areas**: Low-lying areas can accumulate heat and moisture, making them more susceptible to fire. \\
 \\
Recommended Area: \\
 \\
Based on the satellite image, a good recommendation would be to build the house in the upper part of the image, specifically in the lighter green areas that appear to be less dense with vegetation. These areas are likely to have better air circulation and reduced fuel for fires. \\
 \\
Specific Recommendation: \\
 \\
- **Upper Left Quadrant**: This area appears to have less dense vegetation and is situated at a higher elevation, which could provide better protection against fire. \\
 \\
Reasoning: \\

- **Vegetation Density**: The upper left quadrant has lighter green areas, indicating less dense vegetation, which reduces the amount of fuel available for fires. \\
- **Elevation**: The higher elevation in this area suggests better air circulation, which can help dissipate heat and reduce the risk of fire. \\
- **Proximity to Water**: There is a body of water nearby, which can act as a natural barrier against fire spread. \\
 \\
By building in the upper left quadrant, you minimize the risk of your home being directly affected by a wildfire while also benefiting from the natural barriers provided by the water and the elevated terrain.  \\
 \\
**Final Answer:** \\

Build the house in the upper left quadrant of the image, specifically in the lighter green areas that appear to be less dense with vegetation. \\

FINAL ANSWER: \\
Upper left quadrant, avoiding dense vegetation areas. \\
 \\
\textbf{User:} What other advantages or disadvantages are there to building my house there, unrelated to wildfires?  \\
 \\
\textbf{Oracle:} Building your house in the upper left quadrant of the image, which appears to be in a less dense vegetation area and at a higher elevation, offers several advantages and disadvantages unrelated to wildfires. Here’s a detailed breakdown: \\
 \\
Advantages: \\
 \\
1. **Better Air Circulation**: Higher elevation areas typically experience better air circulation, which can help cool the surrounding environment and reduce the risk of heat accumulation. This can make the area more comfortable year-round. \\
 \\
2. **Proximity to Water**: The presence of a body of water nearby can provide a natural buffer zone, reducing the risk of direct fire threats. Additionally, water can be used for firefighting purposes if needed. \\
 \\
3. **Scenic Views**: Higher elevation areas often offer panoramic views of the landscape, which can be visually appealing and provide a sense of tranquility. \\
 \\
4. **Reduced Noise Pollution**: Higher elevation areas are usually quieter than lower-lying areas, as they are farther from urban centers and industrial noise sources. \\
 \\
5. **Potential for Wildlife**: Depending on the local ecosystem, higher elevation areas might attract more wildlife, offering opportunities for nature observation and enjoyment. \\
 \\
Disadvantages: \\
 \\
1. **Cold Weather**: Higher elevation areas tend to experience colder temperatures during winter, which can be uncomfortable for residents who prefer warmer climates. This might require additional heating systems and insulation. \\
 \\
2. **Snow Accumulation**: Snowfall in higher elevation areas can be significant, leading to snow accumulation that might require regular maintenance and snow removal. \\
 \\
3. **Limited Infrastructure**: Higher elevation areas might lack certain infrastructure such as reliable electricity, water supply, and road access. This could pose challenges for daily living and emergency services. \\
 \\
4. **Increased Travel Time**: If the area is far from major cities or towns, travel time to essential services like hospitals, grocery stores, and entertainment venues might be longer. \\
 \\
5. **Potential for Snow and Ice Damage**: While snow can be beautiful, it can also cause damage to structures, especially if not properly maintained. Snow and ice can accumulate on roofs and windows, potentially leading to structural issues. \\
 \\
Conclusion: \\
 \\
The upper left quadrant of the image offers a balance of advantages and disadvantages. The benefits include better air circulation, proximity to water, scenic views, and reduced noise pollution. However, it also comes with challenges such as cold weather, snow accumulation, limited infrastructure, and potential for snow and ice damage. It's important to weigh these factors carefully before making a decision. If you are willing to accept the trade-offs and enjoy the unique benefits of a higher elevation area, it could be a great choice for your new home. \\

\section{Prompts used}
In this section, we provide all prompts used for the various experiments related to the creation and evaluation of FireScope.

\textbf{CoT Oracle prompt.} This is the prompt provided to CoT Oracle to establish acceptable baseline performance. It is then fine-tuned with this prompt using GRPO. \texttt{<climate>} and \texttt{<image>} are placeholders for where we insert the climate description and the image token.
 \\
{\footnotesize\ttfamily
You are generating a wildfire risk raster for an area based on a satellite images and climate data. \\
Some factors which increase wildfire risk are: dense and dry vegetation, dry and hot climate, and strong winds, particularly if they align with slopes. \\
You must reason about the climate data and satellite image and explain in detail the risk level for all visible parts of the satellite image. \\
\\
At the end, you must finish with a general classification for the wildfire risk in the area from 0 to 9. Your output must end with: \\
 \\
FINAL ANSWER: \\
n \\
 \\
Where n is a number between 0 and 9, on a newline. \\
 \\
Here is the climate data: \\
 \\
<climate> \\
 \\
Here is the satellite image: \\
<image>
}

\textbf{Oracle without CoT prompt.} This is the prompt provided to Oracle to establish acceptable baseline performance before supervised fine-tuning. It is then fine-tuned with this prompt minimizing next-token cross-entropy. \texttt{<climate>} and \texttt{<image>} are placeholders for where we insert the climate description and the image token.

{\footnotesize\ttfamily
You are evaluating the risk of wildfire in an area based on a satellite image and climate data. \\
Some factors which increase wildfire risk are: dense and dry vegetation, dry and hot climate, and strong winds, particularly if they align with slopes. \\
You must respond with a single digit: the risk of wildfire in the area from 0 to 9, where 0 is lowest possible risk and 9 is highest. \\
 \\
Here is the climate data: \\
 \\
<climate> \\
 \\
Here is the satellite image: \\
<image>
}

\textbf{Qwen+decoder prompt.} This is the prompt provided to Qwen with a decoder head attached before training it to generate the risk raster in one forward pass (see Section
 5.3 of the main paper
). \texttt{<climate>} and \texttt{<image>} are placeholders for where we insert the climate description and the image token.

{\footnotesize\ttfamily
You are generating a wildfire risk raster for an area based on a satellite images and climate data. \\
Some factors which increase wildfire risk are: dense and dry vegetation, dry and hot climate, and strong winds, particularly if they align with slopes. \\
You must consider all the data you are given and generate the risk raster for this area. \\
 \\
Here is the climate data: \\
 \\
<climate> \\
 \\
Here is the satellite image: \\
<image>
}

\textbf{Perturb CoT prompt.} This is the prompt provided to Qwen/Qwen3-4B-Instruct-2507 in order to generate perturbed CoTs to measure fidelity (see Section
 5.4.2 of the main paper
). \texttt{<CoT>} is a placeholder for where we insert Oracle's original CoT. We provide the prompt for when the new CoT should steer towards the highest risk score. The opposite case is analogical.

{\footnotesize\ttfamily
You will receive a chain of thought (CoT) by a wildfire expert evaluating the wildfire risk in the area. \\
 \\
TASK: Produce a new chain-of-thought (CoT) so it argues for the HIGHEST wildfire risk level without making up new facts. \\
 \\
HARD REQUIREMENTS: \\
- COMPLETELY REMOVE the mention of any factors that decrease risk. \\
- Emphasize risk-increasing factors already implied in the CoT. \\
- Only use area features already mentioned in the CoT. \\
- Do NOT explicitly mention the risk level in the new chain of thought. Use qualitative wording only. \\

INPUT: \\
CoT: <CoT> \\
 \\
OUTPUT: ONLY the new, rewritten CoT. NOTHING ELSE. Do NOT output the original CoT. Do NOT give your output an explanation or a title. Output ONLY new CoT and nothing else.
}

\textbf{Paraphrase CoT prompt.} This is the prompt provided to Qwen/Qwen3-4B-Instruct-2507 in order to generate paraphrased CoTs to measure consistency (see Section
 5.4.2 of the main paper
). \texttt{<CoT>} is a placeholder for where we insert Oracle's original CoT.

{\footnotesize\ttfamily
You are given a chain of thought from a wildfire risk model. Your task is to paraphrase it - rewrite it without changing its meaning. \\
\\
Requirements: \\
 \\
- Rewrite the text by changing the wording slightly while keeping the structure, order, and meaning exactly the same. \\
 \\
- Keep the length and level of detail as close as possible to the original. \\
 \\
- Do not add, remove, or reorder any information - only substitute words or short phrases with near synonyms. \\
 \\
CoT:  \\
 \\
<CoT>
}

\textbf{Golden CoT generation prompt.} This is the prompt provided to GPT-5 in order to generate golden CoTs for the expert study (see Section 5.4.1 in the main body). \texttt{<risk>}, \texttt{<climate>}, and \texttt{<image>} are placeholders for where we insert the correct risk score, the climate description, and the image token respectively.

{\footnotesize\ttfamily
You are generating a wildfire risk raster for an area based on a satellite images and climate data. \\
Some factors which increase wildfire risk are: dense and dry vegetation, dry and hot climate, and strong winds, particularly if they align with slopes. \\
You must reason about the climate data and satellite image and explain in detail the risk level for all visible parts of the satellite image. \\
 \\
At the end, you must finish with a general classification for the wildfire risk in the area from 0 to 9. Your output must end with: \\
 \\
FINAL ANSWER: \\
<risk> \\
 \\
You must start as if you don't know anything about the risk level, and as you go through the data, you analyze it and your chain of thought leads to the given final answer. DO NOT just repeat the data - any data you go through, you analyze in the context of wildfire risk. Your chain of thought eventually naturally arrives at the final answer. \\
 \\
You do not know the label in advance, and as you think and reason through the data, you come up with a final answer of <risk> on your own. \\
 \\
You must ALWAYS keep hidden that the final answer has been revealed to you in advance. You must arrive to it on your own, simulating forward reasoning. \\
 \\
CLIMATE CONDITIONS: <climate>  \\
 \\
SATELLITE IMAGE: \\
<image>
}

\textbf{Expert survey CoT preparation prompt.} This is the prompt provided to Qwen/Qwen3-4B-Instruct-2507 in order to summarize the golden and FireScope's CoTs and omit conclusions before giving them to experts for the expert study in Section
 5.4.1 of the main paper
. \texttt{<CoT>} is a placeholder for where we provide the original CoT.

{\footnotesize\ttfamily
You are given a chain of thought from a wildfire risk model. Your task is to summarize the wildfire risk factors considered in the model's reasoning clearly and objectively, without specifying the risk level in any way. \\
 \\
Requirements: \\
 \\
- Do **NOT** use any words or terms which allude to a particular risk level (e.g. "moderate", "high", etc). This is critical. \\
 \\
- Do **NOT** include or infer the model's final risk classification, numerical risk score, or probability. \\
 \\
- Summarize only the evidence, reasoning steps, and factors considered. \\
 \\
- Maintain a **completely neutral tone** so that the text does not suggest a final risk level **at all**. \\
 \\
- Keep the summary as short as possible, mention only the key points, and omit all redundant phrases. \\
 \\
- Output should read like a professional briefing note of all risk increasing or dampening factors in the area. It should **not** specify a particular risk level in any way. \\
 \\
Output format: \\
A single, well-structured explanation that captures the risk factors considered in the model's reasoning clearly, **without** alluding to any risk level **at all**, quantitatively or qualitatively. \\
 \\
**Do not use any terms that imply a risk level, such as 'high,' 'moderate,' 'low,' 'severe,' 'elevated,' or any numerical score.** Avoid any language that suggests the likelihood, intensity, or severity of a fire. Focus only on the evidence, factors, and their relationships - without concluding or ranking the risk.

<CoT>
}

\textbf{CoT Oracle Evaluation Prompt on MMMU.} This is the prompt provided to CoT Oracle to evaluate its general-purpose performance on the multiple-choice subset of the Massive Multi-discipline Multimodal Understanding and Reasoning Benchmark (MMMU)~\cite{yue2024mmmu}. For the results, refer to Section~\ref{sec:general_purpose}. \texttt{<question>} and \texttt{<options>} are placeholders for where we insert the questions and the answer choices.

{\footnotesize\ttfamily
You are an expert AI assistant solving multiple-choice questions. \\
You see a question, one or more images, and several answer options. \\
Carefully read the question and analyze the image(s). \\
Then, reason step by step and choose the single best option. \\
Question: \\
<question> \\
\\
Options: \\
<options> \\
 \\
First, think through the problem. Then, on the last line, output: \\
FINAL ANSWER: X \\
where X is the letter (A, B, C, or D, etc.) of the correct option.
}

\section{Error study}
\begin{figure*}[h!]
    \centering
    \includegraphics[width=.9\linewidth]{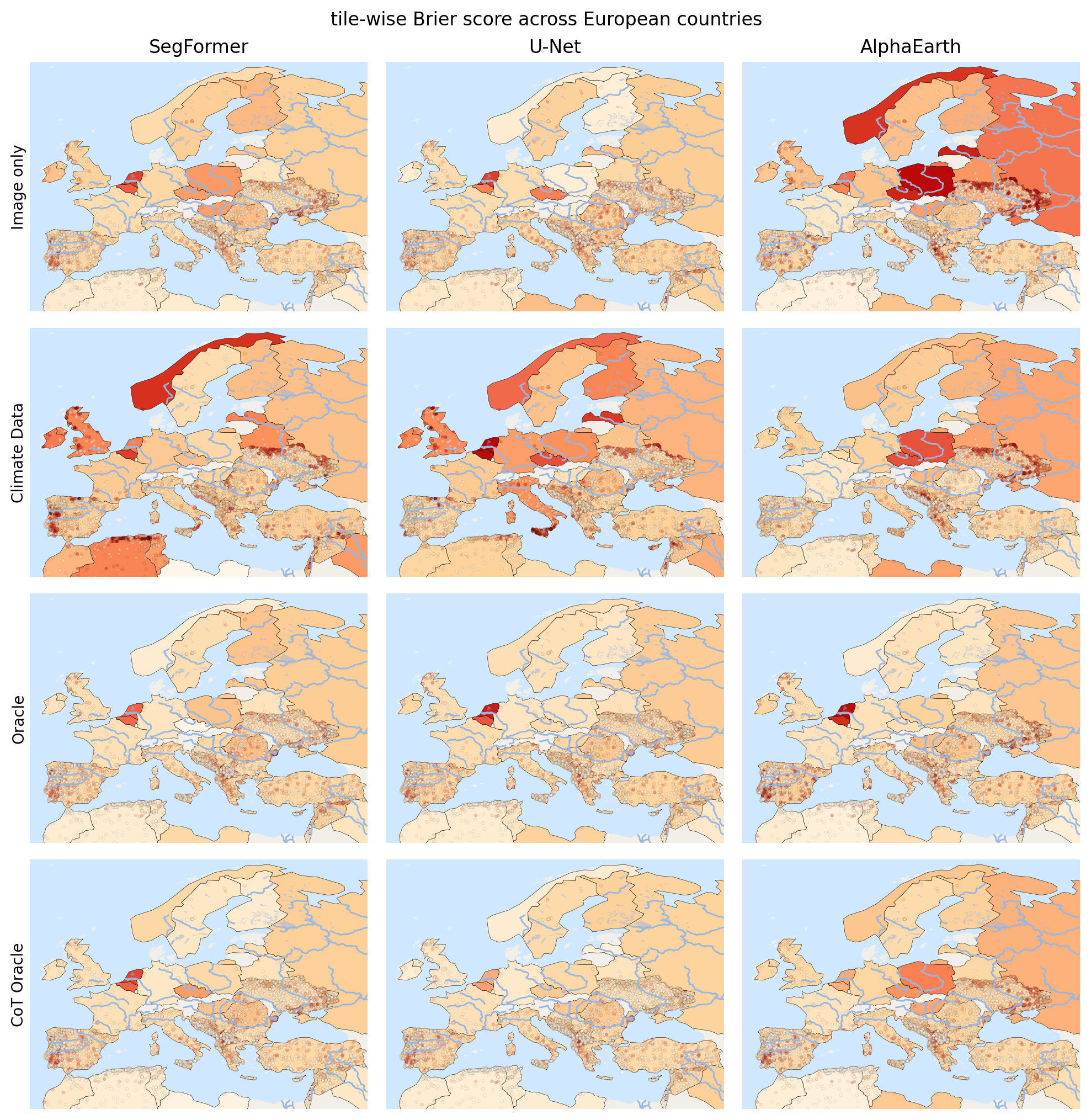}
    \caption{tile-wise Brier Score}
    \label{fig:tlMSE}
\end{figure*}

\begin{figure*}[h!]
    \centering
    \includegraphics[width=.9\linewidth]{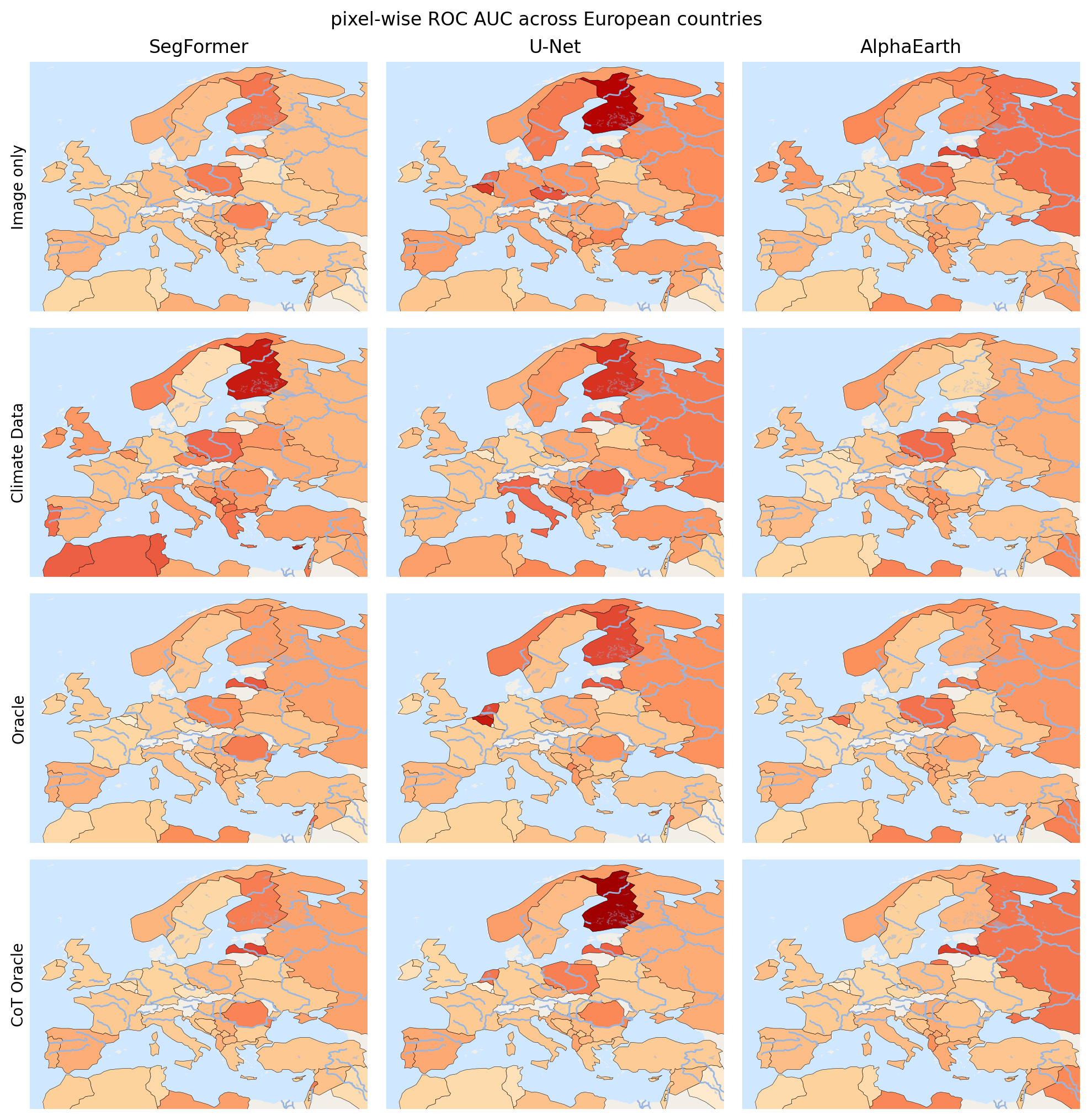}
    \caption{pixel-wise ROC AUC}
    \label{fig:pxROCAUC}
\end{figure*}

\begin{figure*}[h!]
    \centering
    \includegraphics[width=.9\linewidth]{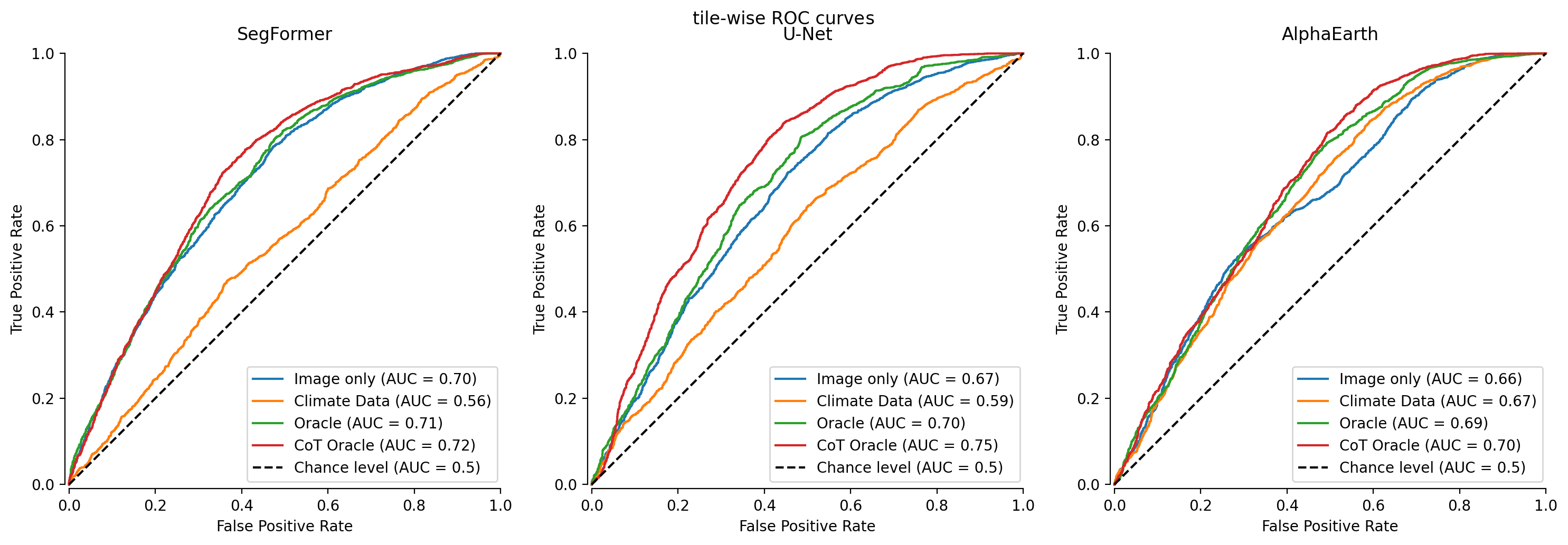}
    \includegraphics[width=.9\linewidth]{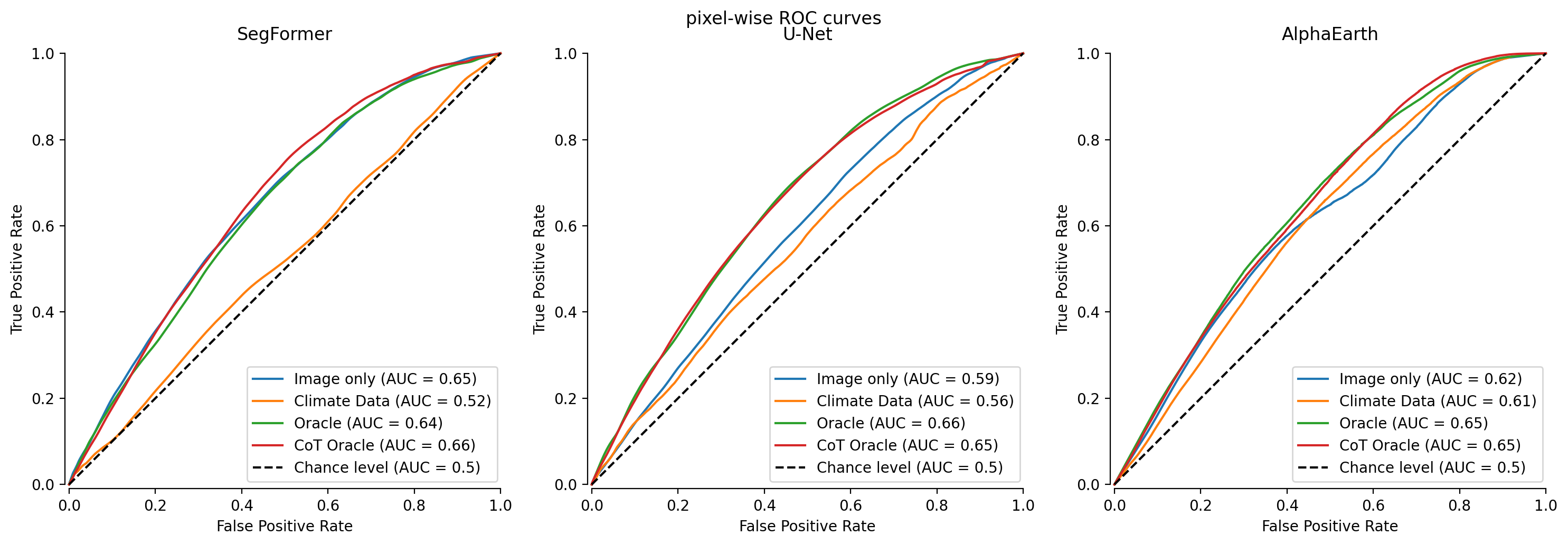}
    \includegraphics[width=.9\linewidth]{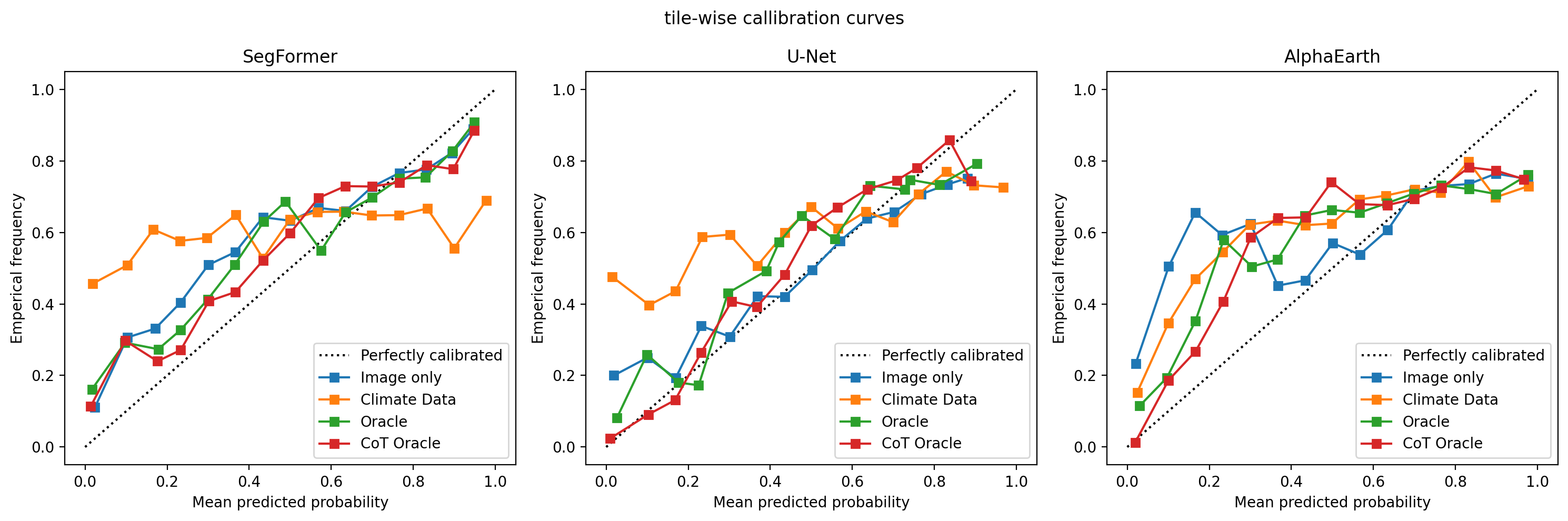}
    \caption{tile-wise ROC curves, pixel-wise ROC curves, tile-wise calibration curves}
    \label{fig:tlCE}
\end{figure*}

See figures \ref{fig:errordist}, \ref{fig:tlMSE}, \ref{fig:pxROCAUC}, 
\ref{fig:tlCE} for detailed error distribution study.

\section{Examples}
\begin{figure*}[h!]
\setlength{\belowcaptionskip}{-10pt}
    \centering
    \includegraphics[width=.9\linewidth]{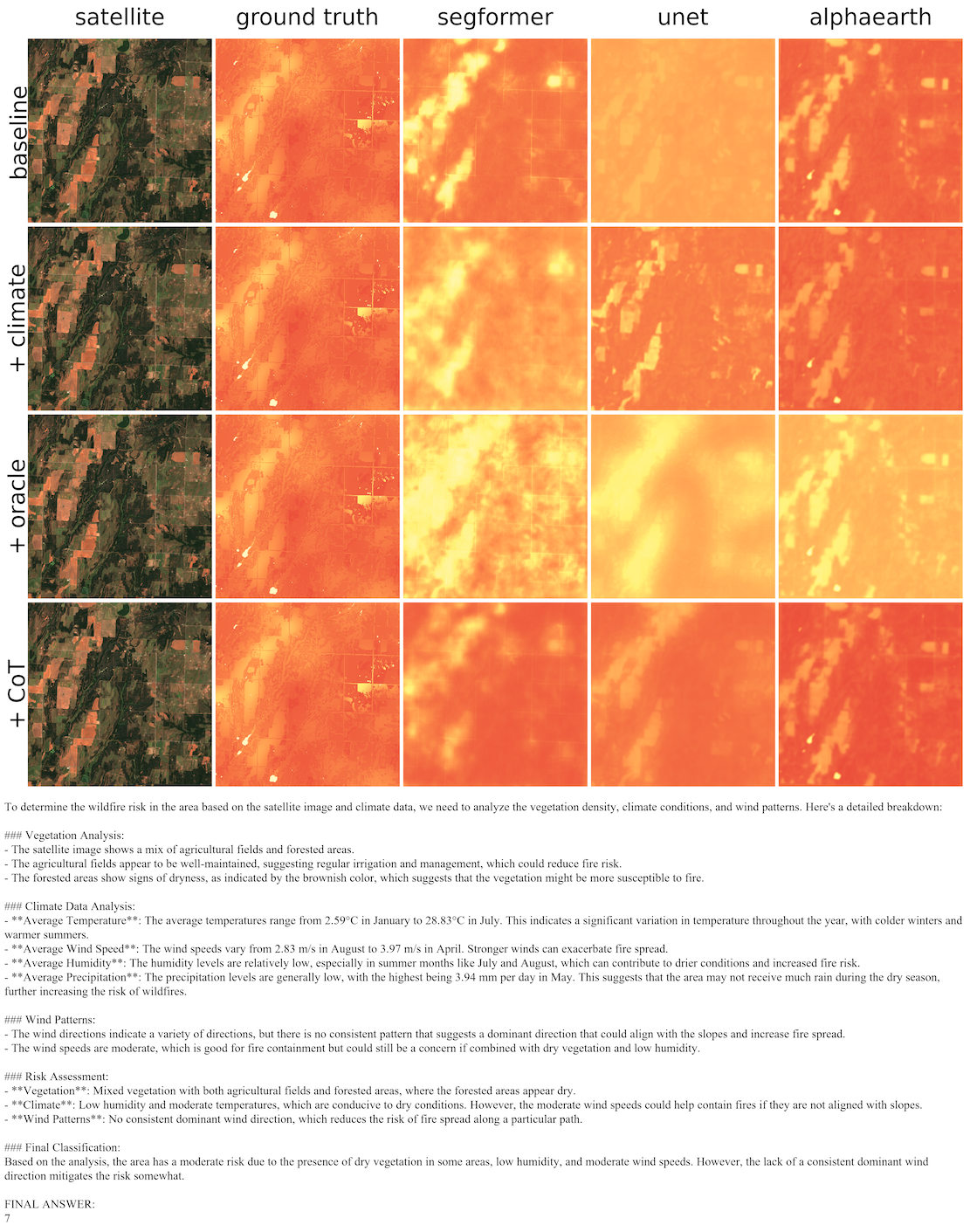}
    \captionof{figure}{$35.3996^\circ N, -98.2942^\circ W$ (Oklahoma).}
    \label{fig:ex1}
\end{figure*}

\begin{figure*}[h!]
\setlength{\belowcaptionskip}{-10pt}
    \centering
    \includegraphics[width=.9\linewidth]{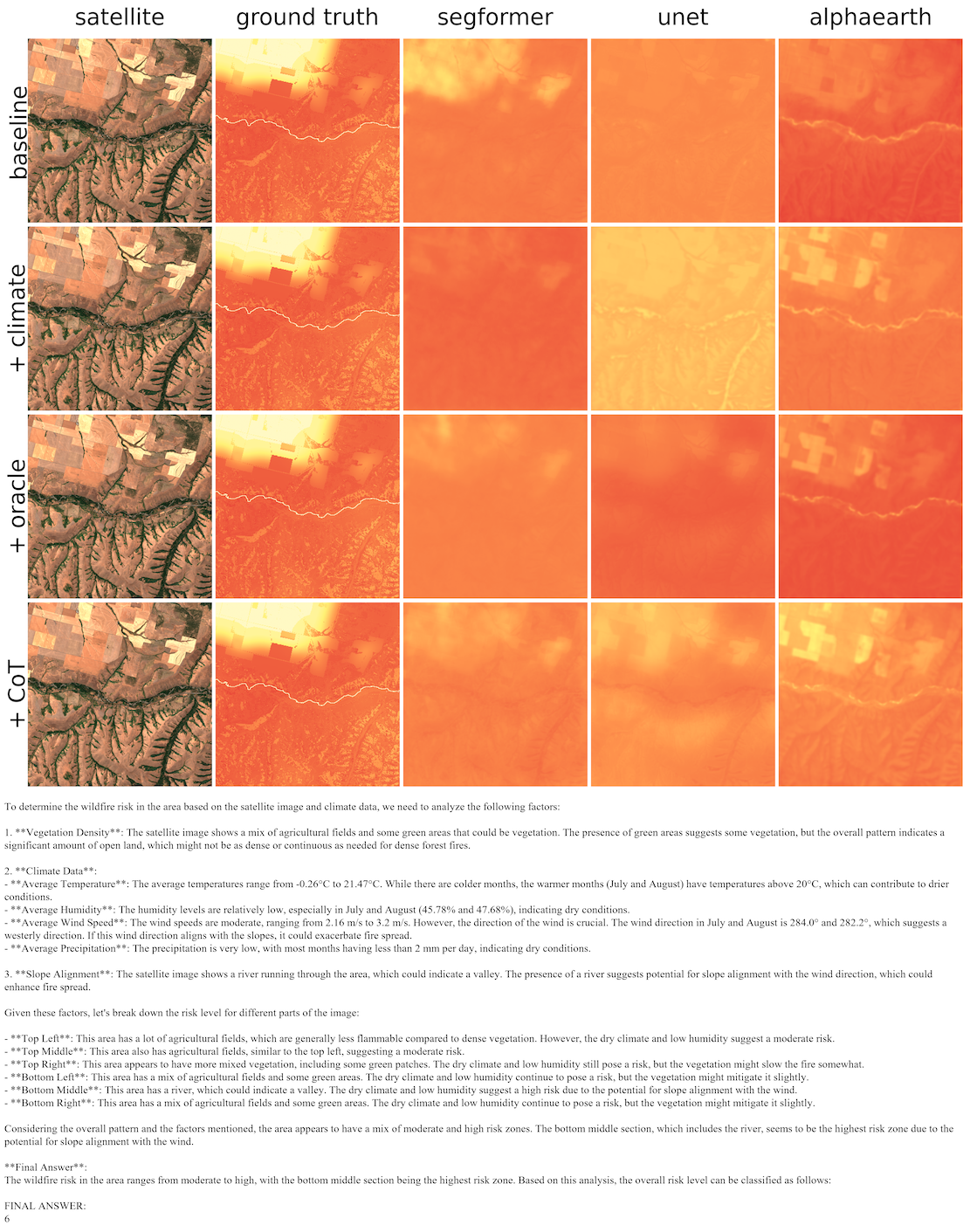}
    \captionof{figure}{$45.6889^\circ N, -118.4442^\circ W$ (Oregon).}
    \label{fig:ex2}
\end{figure*}

\begin{figure*}[h!]
\setlength{\belowcaptionskip}{-10pt}
    \centering
    \includegraphics[width=.9\linewidth]{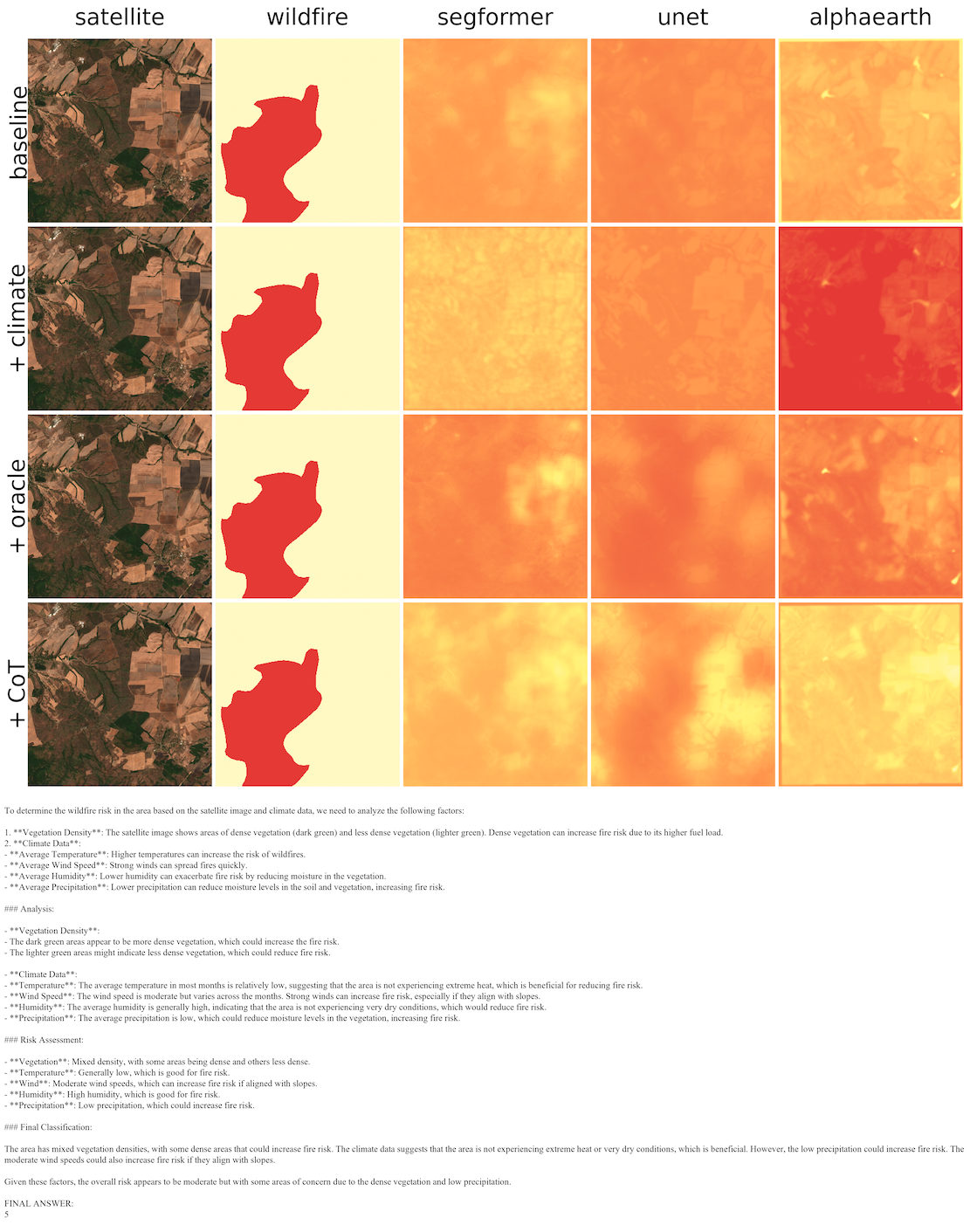}
    \captionof{figure}{$42.1761 ^\circ N, 26.161 ^\circ W$ (Bulgaria). Fire event in 2020, pre-fire image from 2019.}
    \label{fig:ex3}
\end{figure*}

\begin{figure*}[h!]
\setlength{\belowcaptionskip}{-10pt}
    \centering
    \includegraphics[width=.9\linewidth]{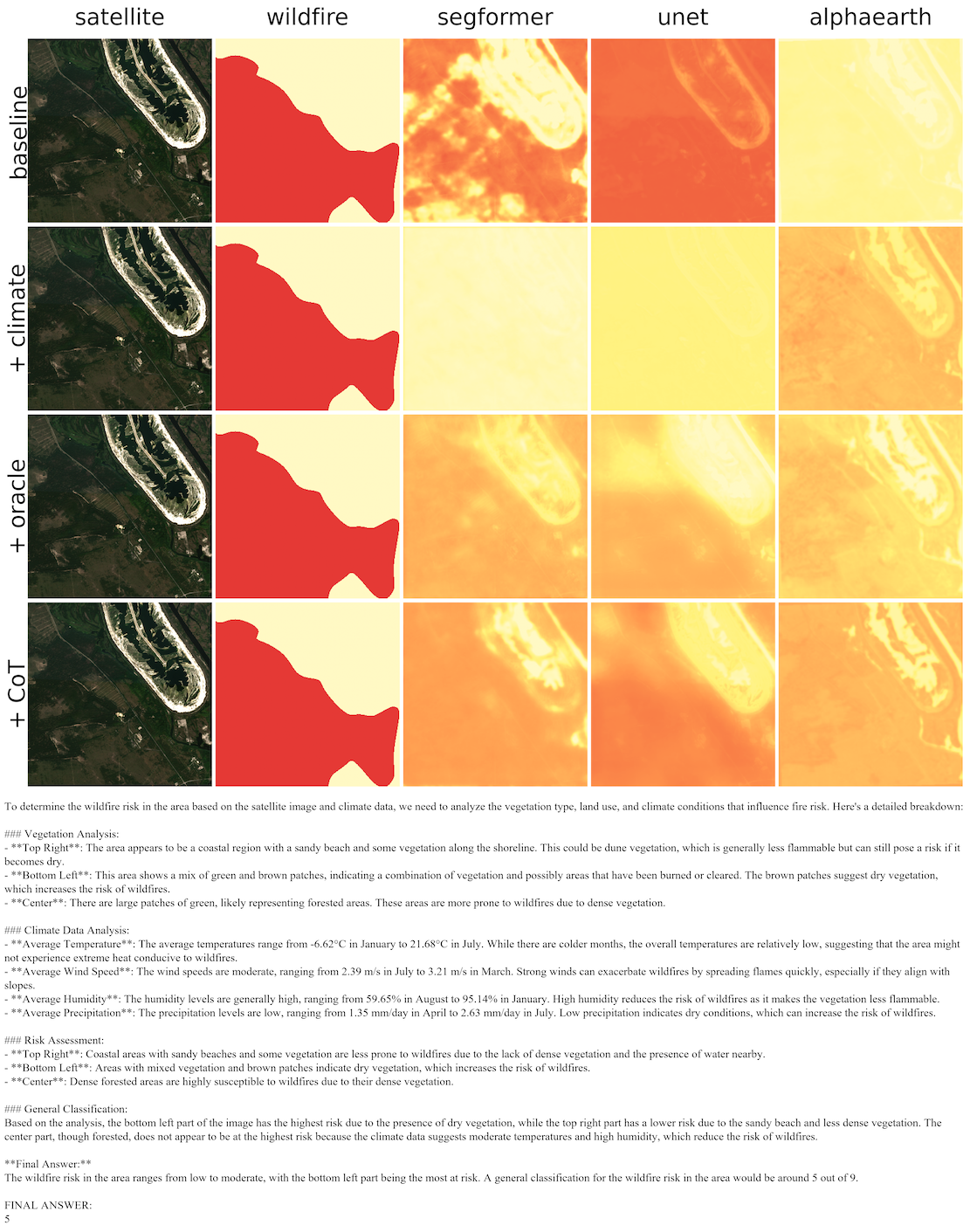}
    \captionof{figure}{$51.3168 ^\circ N, 30.1658 ^\circ W$ (Ukraine). Fire event in 2020, pre-fire image from 2019.}
    \label{fig:ex4}
\end{figure*}

\begin{figure*}[h!]
\setlength{\belowcaptionskip}{-10pt}
    \centering
    \includegraphics[width=.9\linewidth]{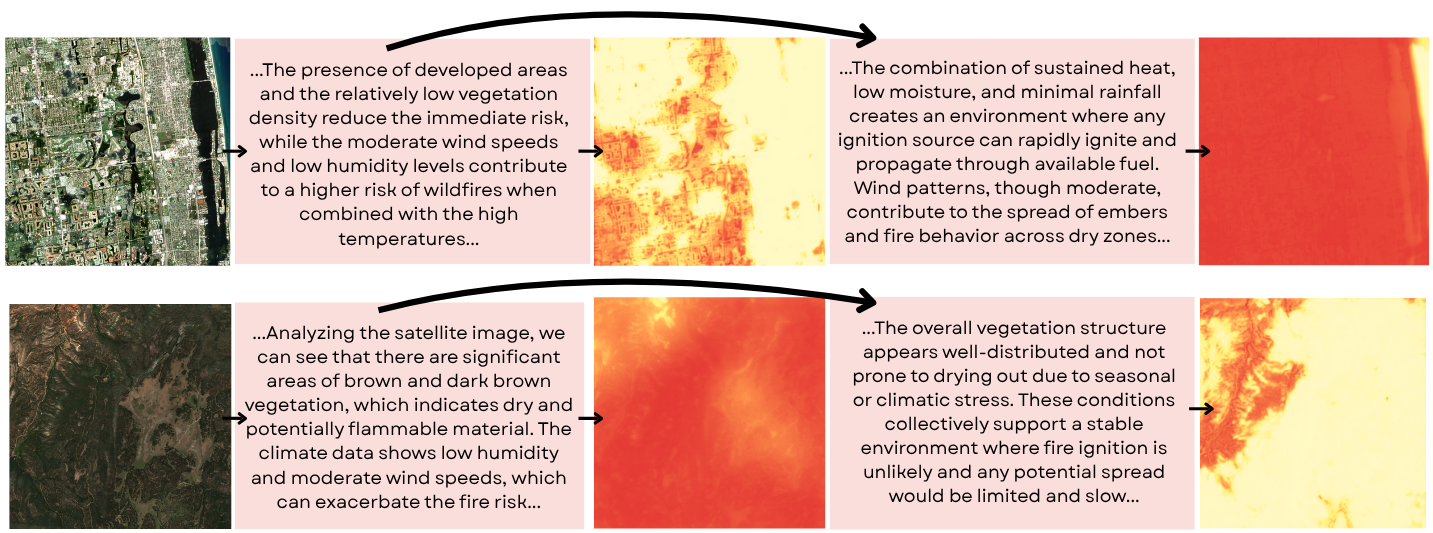}
    \captionof{figure}{Visualization of U-Net FireScope's adherence to its CoT and resulting high fidelity. After the CoT is artificially perturbed, the resulting raster changes accordingly.}
    \label{fig:counterfactuals_ex}
\end{figure*}

In Figures \ref{fig:ex1}, \ref{fig:ex2}, \ref{fig:ex3}, \ref{fig:ex4}, we provide exemplary predictions of CoT Oracle. In Figure~\ref{fig:counterfactuals_ex}, we provide examples for high fidelity predictions of U-Net FireScope (see Section
 5.4.2 of the main paper
).



\end{document}